# Multi-Scale Direction-Aware Network for Infrared Small Target Detection

Jinmiao Zhao, Zelin Shi*, Chuang Yu, Yunpeng Liu, Xinyi Ying and Yimian Dai

***Abstract*—Infrared small target detection faces the problem that it is difficult to effectively separate the background and the target. Existing deep learning-based methods focus on edge and shape features, but ignore the richer structural differences and detailed information embedded in high-frequency components from different directions, thereby failing to fully exploit the value of high-frequency directional features in target perception. To address this limitation, we propose a multi-scale direction-aware network (MSDA-Net), which is the first attempt to integrate the high-frequency directional features of infrared small targets as domain prior knowledge into neural networks. Specifically, to fully mine the high-frequency directional features, on the one hand, a high-frequency direction injection (HFDI) module without trainable parameters is constructed to inject the high-frequency directional information of the original image into the network. On the other hand, a multi-scale direction-aware (MSDA) module is constructed, which promotes the full extraction of local relations at different scales and the full perception of key features in different directions. In addition, considering the characteristics of infrared small targets, we construct a feature aggregation (FA) structure to address target disappearance in high-level feature maps, and a feature calibration fusion (FCF) module to alleviate feature bias during cross-layer feature fusion. Extensive experimental results show that our MSDA-Net achieves state-of-the-art (SOTA) results on multiple public datasets. The code can be available at https://github.com/YuChuang1205/MSDA-Net**

## I. INTRODUCTION

INFRARED imaging is widely used in computer vision tasks such as target detection [1, 2] and image fusion [3, 4] due to its high sensitivity and all-weather operation. Among them, single-frame infrared small target (SIRST) detection is a research hotspot, which focuses on separating small target areas from complex backgrounds [5, 6]. It is widely used in air defense early warning [7, 8], maritime rescue [9, 10] and traffic management [11, 12].

This research was supported by LiaoNing Revitalization Program under Grant no. XLYC2201001. (*Corresponding author: Zelin Shi*)

Jinmiao Zhao and Chuang Yu are with the Key Laboratory of Opto-Electronic Information Processing, Chinese Academy of Sciences, Shenyang 110016, China, also with the Shenyang Institute of Automation, Chinese Academy of Sciences, Shenyang 110016, China, also with the School of Computer Science and Technology, University of Chinese Academy of Sciences, Beijing 100049, China (e-mail: zhaojinmiao@sia.cn; yuchuang@sia.cn).

Zelin Shi and Yunpeng Liu are with the Shenyang Institute of Automation, Chinese Academy of Sciences, Shenyang 110016, China (zlshi@sia.cn; ypliu@sia.cn)

Xinyi Ying is with the College of Electronic Science and Technology, National University of Defense Technology, Changsha, 410073, China (yingxinyi18@nudt.edu.cn).

Yimian Dai is with PCA Lab, VCIP, College of Computer Science, Nankai University, Tianjin, 300350, China. (yimian.dai@gmail.com)

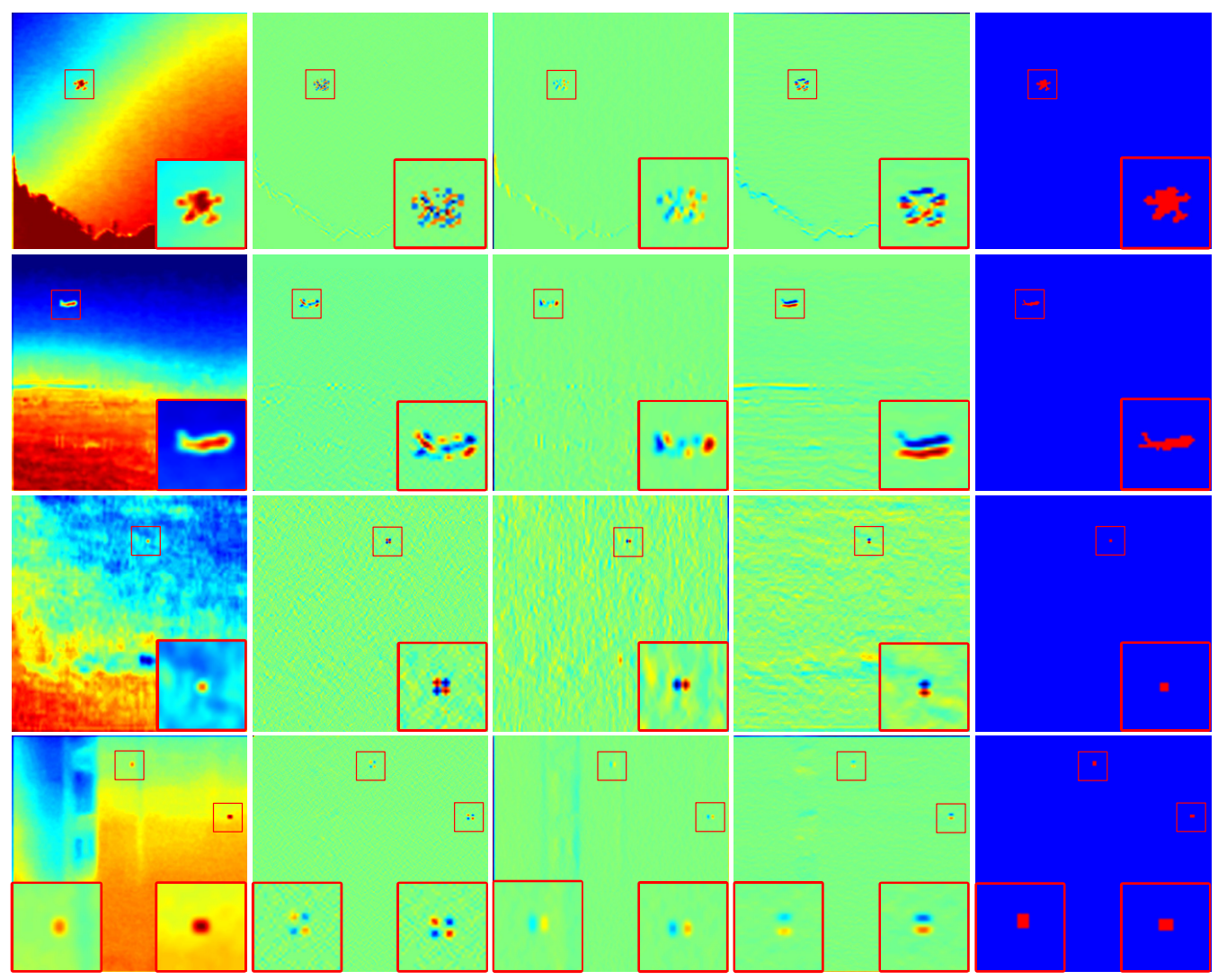

**Fig. 1.** Visualization of high-frequency components in different directions. Each row denotes the original image, diagonal component, horizontal component, vertical component, and true label from left to right. According to the distance difference of the imager, the upper two rows are images with relatively rich shape information of small targets, and the bottom two rows are images with small targets in the shape of spots.

Single-frame infrared small target detection methods can be divided into model-driven methods [14-26] and data-driven methods [13, 32-41, 42-45]. Early studies focused mainly on model-driven methods, which rely mainly on the understanding and modeling of infrared small target images. However, these methods are mostly based on static background or saliency assumptions, which are strongly affected by hyperparameters and have unstable performance. With the development of neural networks, data-driven methods have gradually replaced model-driven methods. The data-driven method inputs infrared images into a deep learning network to learn discriminative features. This type of method can better adapt to a variety of complex environments and reduce the degree of dependence on specific conditions. However, owing to the difficulty in collecting and annotating infrared small target images, purely data-driven methods have limited detection performance [31, 35]. Therefore, we attempt to explore how to integrate domain prior knowledge of infrared small targets into neural networks in a reasonable way, thereby constructing a hybrid method that combines data-driven and model-driven methods.

Recently, high-frequency information, which serves as an important feature carrier of edges, textures, and fine details in images, has attracted increasing attention in data-driven visual perception tasks. Many studies employ frequency-domain modeling to integrate high-frequency information into vision models, enhancing their ability to perceive complex image structures [49-60]. In the SIRST detection tasks, owing to the sensing scenarios, imaging characteristics, and target properties, the key features of targets generally exist in the high-frequency components of the image and often appear in a sparse and weak form. Therefore, effectively modeling and utilizing high-frequency information is essential for enhancing the network's ability to perceive infrared small targets. Some existing methods have implicitly begun to focus on high-frequency components, such as enhancing target edge or shape information to improve detection performance [39, 40]. However, these methods typically focus only on partial high-frequency information, overlooking the richer structural differences and detailed information embedded in high-frequency components from different directions, thus failing to fully exploit the value of directional high-frequency features in target perception. As shown in Fig. 1, focusing on the high-frequency directional component can suppress background interference and highlight target details. Taking the second row as an example, on the one hand, the background clutter in the high-frequency directional component is significantly suppressed. On the other hand, the shape of the aircraft tail in the original image is quite different from that of the true label, and the image obtained after directional filtering can show the tail details. On the basis of the above observations, we believe that directional decoupling of high-frequency information can further explore its value in recognizing small and weak targets. High-frequency features in different directions can highlight the structural and detailed information of small targets from multiple perspectives, thereby enhancing the model's ability to perceive and distinguish target features. Therefore, we first attempt to integrate the high-frequency directional features of infrared small targets as domain prior knowledge into neural networks.

On the basis of the above analysis, we propose an innovative multi-scale direction-aware network (MSDA-Net). Specifically, to fully mine the high-frequency directional features, on the one hand, we propose a high-frequency direction injection (HFDI) module without trainable parameters in the initial part of the network. This module makes full use of the multi-directional high-frequency components of the original image to help guide the network to pay attention to detailed information such as target edges and textures. On the other hand, we propose a multi-directional feature awareness (MDFA) module on the premise of imitating human visual characteristics. On this basis, we integrate the multi-scale local relation learning (MLRL) module and further construct a multi-scale direction-aware (MSDA) module, which promotes the full extraction of local relations at different scales and the full perception of key features in different directions. In addition, owing to the characteristics of this task, the target features in high-level feature maps are easily overwhelmed by redundant background features as the network layers deepen, and even slight feature bias can also significantly impact on detection performance. On the one hand, to solve the problem that target features are overwhelmed in deep feature maps, we propose a simple and effective feature aggregation (FA) structure, which employs max pooling operation to reasonably aggregate multi-scale low-level features to supplement the target features in the high-level feature maps. On the other hand, to alleviate feature bias when multi-level feature maps are fused, we construct a lightweight feature calibration fusion (FCF) module, which allows high-level features and low-level features to be calibrated before fusion, thereby reducing cross-layer feature bias and enabling more accurate fusion. The contributions of this study can be summarized as follows:

1) A novel MSDA-Net for SIRST detection is proposed, which is the first to integrate the high-frequency directional features of infrared small targets as domain prior knowledge into neural networks.

2) To fully mine the high-frequency directional features, we innovatively construct a HFDI module to guide the network to focus on detailed information, and a MSDA module to facilitate the full extraction of local region relations at different scales and the full perception of key features in different directions.

3) Considering the characteristics of infrared small targets, we propose a FA structure to address target disappearance in high-level feature maps, and a FCF module to alleviate feature bias during cross-layer feature fusion.

4) Extensive experiments on the multiple public datasets demonstrate that our MSDA-Net achieves SOTA performance and exhibits strong generalizability.

## II. Related Work

This section provides a brief review of representative methods for the SIRST task, including both non-deep learning-based SIRST detection methods and deep learning-based SIRST detection methods, along with an overview of the high-frequency information in visual perception.

### *A. Non-deep learning-based SIRST detection methods*

Non-deep learning-based SIRST detection methods can be categorized into three main categories: background suppression-based methods [14-18], human visual system-based methods [19-22] and image data structure-based methods [23-26]. Background suppression-based methods can be further divided into spatial domain filtering methods [14-16] and transform domain filtering methods [17-18]. The former offers high computational efficiency but suffers from poor robustness in complex backgrounds, whereas the latter can enhance target responses but incurs a high computational cost. Human visual system-based methods [19-22] simulate the information processing process of the human eye when discovering and locking targets. This type of method performs well in suppressing large areas of bright background, but is prone to false detection in the presence of strong noise points. Image data structure-based methods [23-26] mainly utilize the non-local self-similarity of the background and the sparse characteristics of the target to transform SIRST detection tasks into low-rank and sparse matrix separation problems. This type of method can better separate the background and foreground, but it is still sensitive to strong edge backgrounds and has a high computational cost.

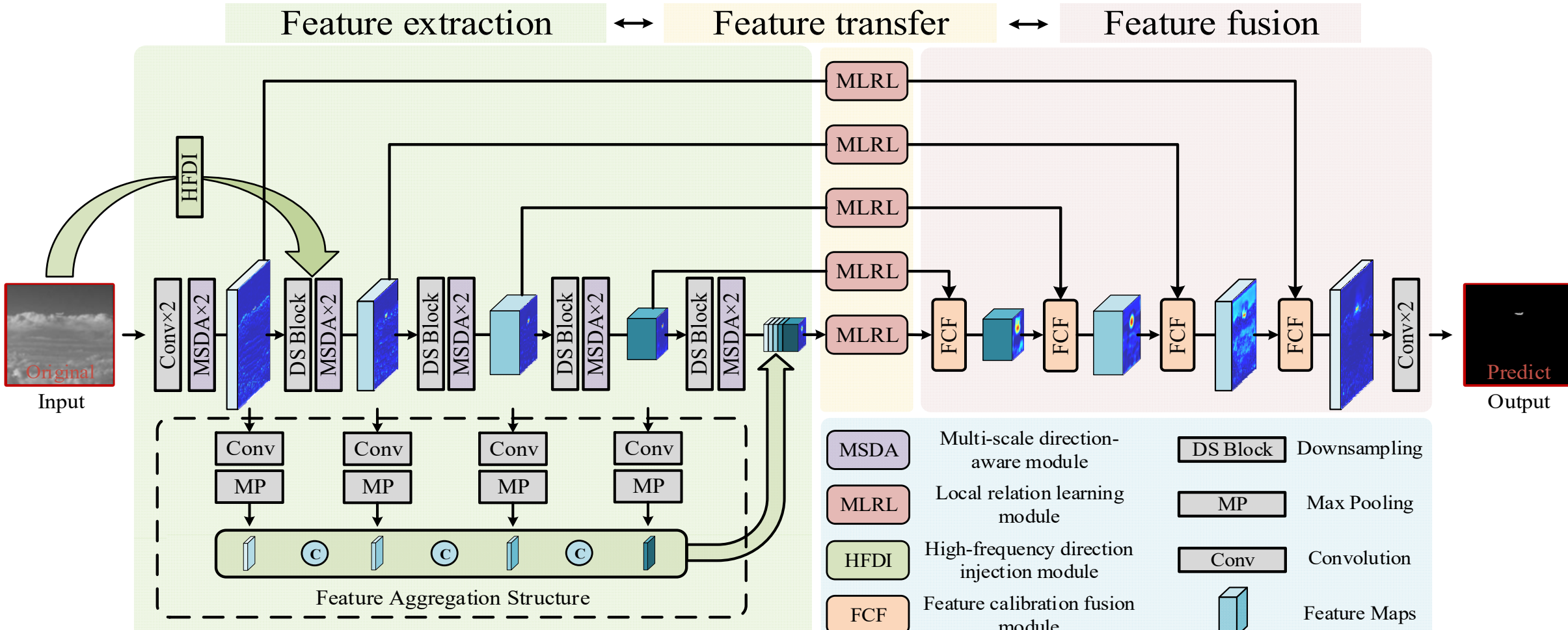

**Fig. 2.** Overall structure of MSDA-Net. MSDA-Net consists of three parts: feature extraction, feature transfer and feature fusion. The feature extraction part is used to fully extract the features of infrared small targets. The feature transfer part is used to further extract the relationships between local areas at different scales. The feature fusion part is used to achieve effective and accurate fusion between cross-layer feature maps.

*B. Deep learning-based SIRST detection methods*

The core idea of deep learning is to achieve pattern recognition and representation learning tasks by constructing and training deep neural networks [27-30]. In recent years, with the development of deep learning, deep learning-based SIRST detection methods [13, 32-41] have gradually surpassed non-deep learning methods. Deep learning-based methods can be mainly divided into exploration of the network structure, utilization of contrast information, and emphasis on edge shape information. Among them, the exploration of network structures has focused mainly on the improvements in backbone architectures to enhance the extraction of discriminative features. Most existing methods are based on modifications to the excellent U-Net [31] framework, including pure CNN-based networks such as the asymmetric context module (ACM) [32], the attention guided pyramid context network (AGPCNet) [33], the densely nested attention network (DNA-Net) [34], the U-Net in U-Net (UIU-Net) [35], and the multi-scale head to the plain UNet (MSHNet) [41], as well as networks that integrate Vision Transformer [42] architectures, such as the multi-level TransUNet (MTU-Net) [43], the spatial channel cross transformer network (SCTransNet) [44] and the shallow-deep synergistic detection network (SDS-Net) [45], to enhance long-range dependency modeling capabilities. Moreover, some studies have designed structurally constrained networks grounded in optimization theory, such as PRCANet [46]. The utilization of contrast information is mainly inspired by the local contrast idea of non-deep learning methods. The attention-based local contrast network (ALCNet) [36], the multi-scale local contrast learning network (MLCL-Net) [37] and the attention-based local contrast learning network (ALCL-Net) [38] are proposed one after another. Recently, some researchers have devoted themselves to making full use of the edge shape information of small targets, including the infrared shape network (ISNet) [39], the gradient-guided learning network (GGL-Net) [40], the shape-biased representation network (SRNet) [13], the multi-branch mutual-guiding learning network (MMLNet) [47] and the infrared low-level network (ILNet) [48]. In contrast, this study aims to explore a new idea of converting domain prior knowledge of infrared small target images into feature representations and injecting them into the network. Our MSDA-Net can fully extract local relations at different scales and fully perceive key features in different directions, thereby achieving the refined detection of infrared small targets.

*C. High-frequency information in visual perception*

Recently, frequency domain modeling has received extensive attention in visual perception tasks. Among them, high-frequency information is an important component that includes edges, textures, and fine details in images. Existing studies have explicitly modeled high-frequency information in images across a variety of tasks, including image compression [49], super-resolution reconstruction [50, 51, 52], image classification [53, 54], semantic segmentation [55, 56], change detection [57, 58], and object detection [59, 60]. Specifically, in image compression and reconstruction, HL-RS CompNet [49] and DDOM [50] alleviate issues such as detail loss and structural distortion under high compression ratios by introducing high-frequency reconstruction mechanisms. In image classification and semantic segmentation, RSFJR [53] employs frequency filtering to extract high-frequency information, enhancing boundary and object perception, whereas the adaptive frequency enhancement network (AFENet) [55] incorporates high-frequency components during feature fusion to improve the discrimination of fine textures and target boundaries. In object detection and change detection, the multiscale spatial-frequency domain network (MSFN) [59] uses multi-scale high-frequency information to guide the network's attention to focus on angle changes, thereby improving the detection accuracy of the model in complex backgrounds. Meanwhile, the joint frequency-spatial domain network (JFSDNet) [57] enhances sensitivity to small structural changes through frequency channel reweighting, thereby improving the model's sensitivity to small change areas and enhancing its robustness. The above studies demonstrate that high-frequency modeling is generally effective in enhancing the representation of texture details. Considering that infrared small targets typically exist in the high-frequency regions of images in a weak and complex form, effective modeling of high-frequency features is crucial for fine detection. However, existing methods focus mostly on features such as edges and shapes, but overlook the richer structural differences and detailed information embedded in high-frequency components from different directions. Therefore, this paper explores incorporating high-frequency directional features as domain priors into deep networks to enhance the model's capability in perceiving small targets.

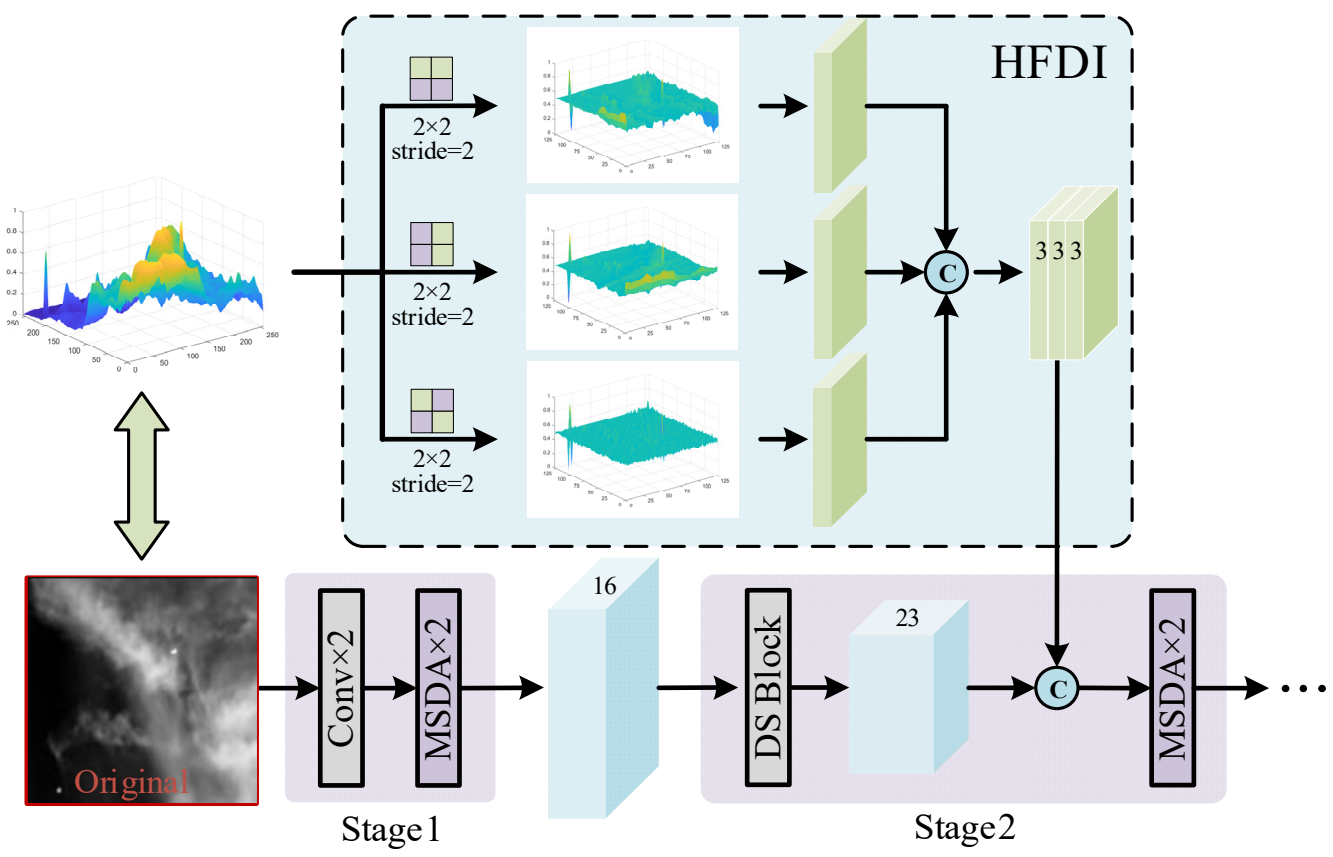

**Fig. 3.** High-frequency direction injection module.

## III. Method

### *A. Multi-Scale Direction-Aware Network*

As shown in Fig. 2, the proposed MSDA-Net consists of three parts: feature extraction, feature transfer and feature fusion. For the feature extraction part, the input image is processed through both an auxiliary branch and a main branch. The auxiliary branch consists of a HFDI module, which extracts multi-directional high-frequency information from the original image and injects it into the main branch at the early stage to guide the model in focusing on critical details. The main branch consists of five stages: the first stage with two convolutions and two MSDA modules, and the remaining four stages each with a downsampling module and two MSDA modules. As a core component, the MSDA module promotes the full extraction of local relations at different scales and the full perception of key features in different directions. It is composed of three sub-modules: a MLRL module, a MDFA module, and a SE attention module [58]. The structure of the downsampling module is consistent with the Block structure in our previous works [37, 40]. Additionally, we propose a FA structure that downsamples the feature maps from the first four stages via max pooling to match the resolution of the final stage. These feature maps are then concatenated along the channel dimension and compressed through a 1×1 convolution layer. This design helps alleviate the disappearance of small targets caused by network deepening and allows the feature map from the fifth stage to obtain accurate spatial information while having strong semantic information. For the feature transfer part, we use the MLRL module to further extract the relationships between local areas at different scales. This module has the same structure as the MLRL module in the MSDA module. For the feature fusion part, the FCF module is built and used to achieve effective fusion of feature maps with different resolutions. This module effectively alleviates the target feature bias in multi-level feature fusion by guiding high-level features to learn a calibration map relative to low-level features.

### *B. High-frequency direction injection module*

The key information required for SIRST detection, such as the target edge, shape and textures, is reflected in the high-frequency components of the image. To highlight fine-grained target details and suppress background clutter at the early stage of the network, we construct a HFDI module. The structure of the HFDI is shown in Fig. 3. Notably, the proposed HFDI module is a processing module that is based on prior knowledge and has no learning parameters. To effectively model directional high-frequency features in infrared images, we design three fixed high-pass convolution kernels corresponding to the LH, HL and HH components in the two-dimensional Haar wavelet transform. These kernels are used to extract the high-frequency responses in the horizontal, vertical and diagonal directions, respectively. As a classic signal processing tool, the Haar wavelet is known for its strong directional sensitivity, powerful edge enhancement capability and high computational efficiency, making it particularly suitable for highlighting local variations of small targets in infrared images. Specifically, we channel-stitch the three high-frequency components in the horizontal, vertical, and diagonal directions of the original image with the feature map generated after the first stage and downsampling. The spliced feature map is used as the subsequent input of the second stage. The convolution kernels for extracting high-frequency components in the horizontal, vertical, and diagonal directions are $\begin{bmatrix} 0.5 & -0.5 \\ 0.5 & -0.5 \end{bmatrix}$, $\begin{bmatrix} 0.5 & 0.5 \\ -0.5 & -0.5 \end{bmatrix}$, and $\begin{bmatrix} 0.5 & -0.5 \\ -0.5 & 0.5 \end{bmatrix}$, respectively, and the stride is 2.

The HFDI module explicitly injects directional high-frequency information into the early feature representation, providing structural and detail information from multiple perspectives of the original image. This effectively suppresses background clutter, highlights potential small target regions, and enhances the network's understanding of target structures and background context. In the datasets, the small targets are divided into two situations: 1. When the imager is closer to the target, the obtained small target shape information is richer. 2. When the imager is far from the target, owing to the optical point diffusion characteristics of the thermal imaging system and long-distance imaging, the small target appears spotty, and the structure and shape information are weak. For the two different situations, the original images are passed through the above-mentioned high-pass filters in the horizontal, vertical, and diagonal directions. From Fig. 1, the cluttered backgrounds in both types of infrared small target images are significantly suppressed by directional filtering, and the edge details are effectively enhanced. Furthermore, to validate the rationality and effectiveness of the selected directional operator, we design a detailed comparative experiment and provide a systematic analysis in Section IV-E.

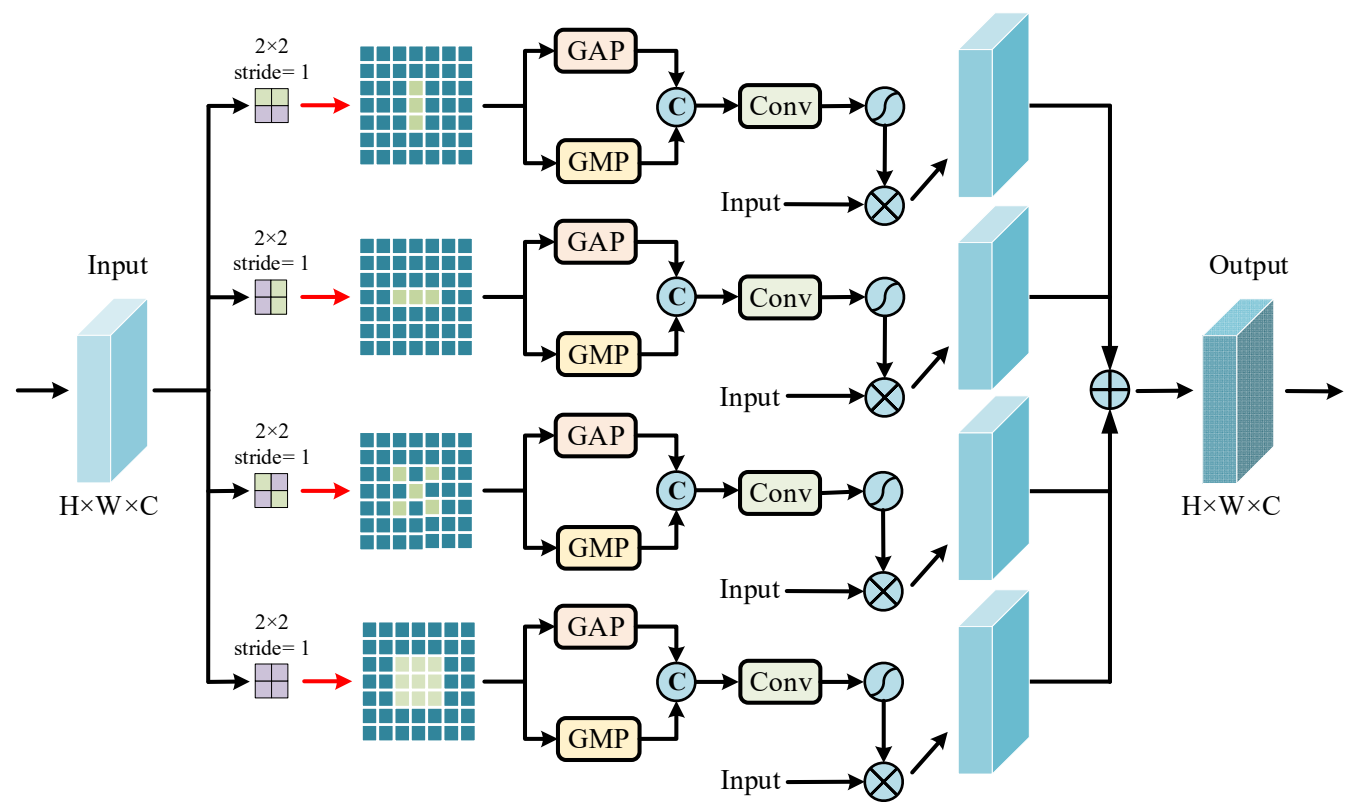

**Fig. 4.** Multi-directional feature awareness module.

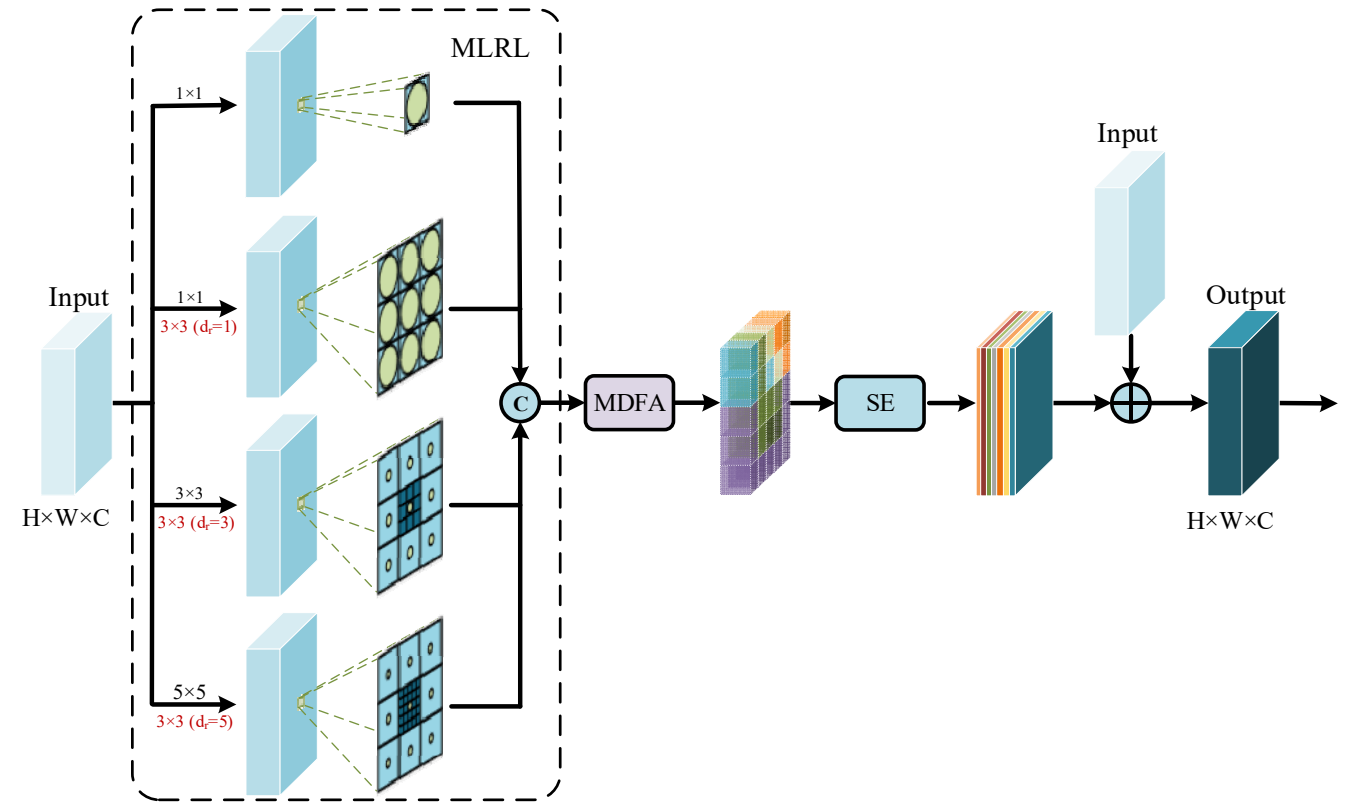

**Fig. 5.** Multi-scale direction-aware module. "$d_r$" denotes the dilation rate.

### *C. Multi-directional feature awareness module*

In the feature extraction part, to fully emphasize the network's ability to extract high-frequency directional features while also preserving low-frequency global features (zero directional), we construct an MDFA module. Following the same idea as the HFDI module design, we adopt the two-dimensional Haar wavelet transform operators to extract multi-directional information. From Fig. 4, the first input feature map is passed through four filters, namely, horizontal, vertical, diagonal and low frequency, to obtain the components in each direction of the feature map. Secondly, the obtained components are spliced in the channel dimension after global average pooling and global maximum pooling to better locate the position of the small target. Thirdly, the obtained fine components are used to apply attention to the original feature map, thereby obtaining the feature maps that focus on positions in each direction. Finally, the obtained feature maps in four directions are fused to obtain the final output. The formula for this structure is expressed as follows:

$$F_{out} = F_{H_h} \oplus F_{H_v} \oplus F_{H_d} \oplus F_L \tag{1}$$

$$F_{H_h} = S\Big( f_c(G_{AP}(f_{H_h}(F_{in})) \oplus_c (G_{MP}(f_{H_h}(F_{in})))\Big) \otimes F_{in} \tag{2}$$

$$F_{H_v} = S\Big( f_c(G_{AP}(f_{H_v}(F_{in})) \oplus_c (G_{MP}(f_{H_v}(F_{in})))\Big) \otimes F_{in} \tag{3}$$

$$F_{H_d} = S\Big( f_c(G_{AP}(f_{H_d}(F_{in})) \oplus_c (G_{MP}(f_{H_d}(F_{in})))\Big) \otimes F_{in} \tag{4}$$

$$F_L = S\Big( f_c(G_{AP}(f_L(F_{in})) \oplus_c (G_{MP}(f_L(F_{in})))\Big) \otimes F_{in} \tag{5}$$

where $F_{in}$ and $F_{out}$ denote the input and output, respectively. $f_{H_h}$, $f_{H_v}$, $f_{H_d}$ and $f_L$ denote convolutions with convolution kernels $\begin{bmatrix} 0.5 & -0.5 \\ 0.5 & -0.5 \end{bmatrix}$, $\begin{bmatrix} 0.5 & 0.5 \\ -0.5 & -0.5 \end{bmatrix}$, $\begin{bmatrix} 0.5 & -0.5 \\ -0.5 & 0.5 \end{bmatrix}$, $\begin{bmatrix} 0.5 & 0.5 \\ 0.5 & 0.5 \end{bmatrix}$ and a stride of 1, respectively. $G_{AP}$ denotes the global average pooling. $G_{MP}$ denotes the global maximum pooling. $f_c$ and $S$ denote the convolution and sigmoid operations, respectively. $\oplus_c$ denotes the splicing of the channel dimensions. $\otimes$ and $\oplus$ denote element-wise multiplication and element-wise addition.

This module focuses on both high-frequency directional features and low-frequency overall features in the spatial dimension. Specifically, applying attention to high-frequency information in multiple directions is beneficial for precisely extracting the structure and position of infrared small targets. Despite the presence of complex backgrounds and interfering regions with brightness levels equal to or even higher than that of the target, background clutter typically varies slowly within local neighborhoods. Therefore, applying attention to high-frequency information in different directions can effectively suppress background clutter while enhancing discriminative features for the SIRST detection tasks. From the perspective of the visual system, this structure simulates the suppression of commonly consistent responses and the emphasis on outliers that appear in each directional feature. Meanwhile, applying attention to low-frequency information enables the network to capture the overall structure and high-level semantic information of the image, thereby enhancing its ability to extract global features. By paying attention to low-frequency information, the network can avoid being misled by local noise or fragmented details, thereby identifying key regions more accurately and robustly. From the perspective of the visual system, this structure simulates how the visual system captures global contextual information to form a coherent understanding of the overall scene.

### *D. Multi-Scale Direction-Aware module*

In infrared images, the temperature difference of an object creates a contrast difference, which is crucial for locating and identifying targets [23]. Therefore, building upon MDFA, to fully model directional features while effectively utilizing the relationships between local areas in infrared images, we propose a MSDA module. Its structure is shown in Fig. 5. As a fundamental component of the feature extraction part, this module consists of three sub-modules: the MLRL module, the MDFA module, and the SE attention module. Specifically, the MSDA module can be expressed as:

$$F_{output} = A_s(A_d(E_r(F_{input}))) \oplus_{res} F_{input} \tag{6}$$

where $E_r$ denotes the MLRL module. $A_d$ denotes the MDFA module. $A_s$ denotes the SE attention module. $\oplus_{res}$ denotes the residual connection.

First, on the basis of our previous research [37], we construct the MLRL module, which uses a combination of multi-scale convolution and dilated convolution to learn the relationships between local areas at different scales. Taking the bottom branch as an example, first, local area patch features with an area of 5×5 pixels can be extracted through a 5×5 convolution. Next, the relationships between the corresponding 5×5 area patch and its 8 adjacent area patches can be further extracted through a 3×3 dilated convolution with a dilation rate of 5. Subsequently, we use a 1×1 convolution to fuse the multi-scale relation feature maps after channel splicing and achieve channel dimensionality reduction. The formulas can be expressed as:

$$F_{b_0} = Conv_{1\times1}(F_{input}) \tag{7}$$

$$F_{b_1} = DConv_1(Conv_{1\times1}(F_{input})) \tag{8}$$

$$F_{b_2} = DConv_3(Conv_{3\times3}(F_{input})) \tag{9}$$

$$F_{b_3} = DConv_5(Conv_{5\times5}(F_{input})) \tag{10}$$

$$E_r(F_{input}) = Conv_{1\times1}(F_{b_0} \oplus_c F_{b_1} \oplus_c F_{b_2} \oplus_c F_{b_3}) \tag{11}$$

where $F_{b_0}$, $F_{b_1}$, $F_{b_2}$, and $F_{b_3}$ denote the corresponding feature maps after each branch. $DConv_n$ denotes a dilated convolution with a convolution kernel of 3×3, and its subscript $n$ denotes the dilation rate. On the one hand, the use of the MLRL module can enhance the network's adaptability to small targets of different scales, thereby capturing detailed information of the target more comprehensively. On the other hand, it helps guide the network to consider local area relations, allowing the network to better understand and capture the relationships between small targets and their surrounding environments, such as contrast differences. Subsequently, the features extracted by the MLRL module are passed to the MDFA module to guide the network to focus on both high-frequency directional features and low-frequency global features in the scale space. Then, the feature maps are further processed with channel-level attention weighting through the SE module, which dynamically adjusts the importance of each channel. Finally, the SE module's output is fused with the original input through a residual connection, enabling efficient information transmission and enhanced feature representation.

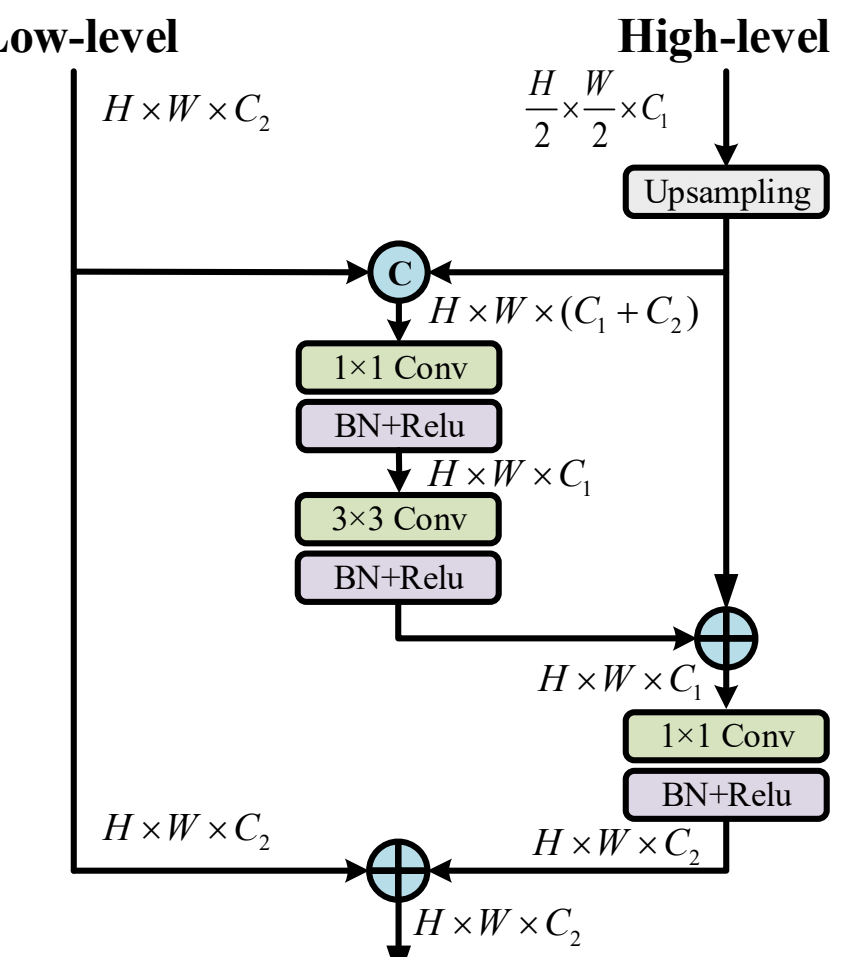

**Fig. 6.** The feature calibration fusion module.

The MSDA module jointly models scale relationships and direction information, thereby effectively promoting the full extraction of local region relationships at different scales and the full perception of key features in different directions.

### *E. Feature Calibration Fusion module*

The target size in this task is small, and even slight feature bias have a significant impact on the detection performance. To address the problem in cross-layer feature fusion, we propose a FCF module. As shown in Fig. 6, the core idea of the FCF module is to perform feature-domain calibration before fusion: low-level features guide high-level features to learn a relative calibration map, thereby reducing cross-layer feature bias and improving fusion quality.

Specifically, given a low-level feature map $F_L$ and a high-level feature map $F_H$, the size of $F_L$ is $H \times W \times C_2$, and the size of $F_H$ is $H/2 \times W/2 \times C_1$. First, $F_H$ is upsampled and spliced with $F_L$ in the channel dimension. Secondly, the channel dimension compression of the spliced feature map is achieved through a 1×1 convolution so that the output is the same as the channel dimension of $F_H$. The compressed feature map is passed through a 3×3 convolution to learn the calibration map of the high-level feature map $F_H$ relative to the low-level feature map $F_L$. Then, the feature map $F_H$ and the learned relative calibration map are added and passed through a 1×1 convolution to calibrate it with $F_L$ again in the channel dimension. Finally, the calibrated high-level feature map and the low-level feature map are fused. Notably, all the FCF modules account for only 3.25% of the total network parameters, which is very lightweight. They significantly enhance feature fusion with minimal overhead.

Interestingly, we find that the low-level branch of the FCF module resembles a residual structure from another perspective. It has been verified in ResNet [62] that the residual structure can learn differential features. This further verified that the proposed FCF module can learn the refine calibration map between high-level feature maps relative to low-level feature maps, thereby achieving effective and precise fusion between cross-layer feature maps.

## IV. Experiment

### *A. Datasets*

We use three publicly available datasets: NUDT-SIRST [34], SIRST [32], and IRSTD-1k [39] to evaluate the performance of the methods. For the SIRST and IRSTD-1k datasets, we uniformly resize the images to 512×512 pixels. For the NUDT-SIRST dataset, we uniformly resize the images to 256×256 pixels. Additionally, we use an infrared video dataset (IRSatVideo-LEO [1]) to evaluate the performance of the proposed method on multi-frame infrared small target detection tasks, thereby validating its potential for temporal generalization. The details are as follows:

*1) NUDT-SIRST dataset*. It is a synthetic dataset with five main background scenes: city, field, highlight, ocean and cloud. Approximately 37% of images contain at least two objects, and 32% of objects are located outside the top 10% of the image brightness values. For this dataset, we adopt two division rules: 663: 664 and 1027: 300.

TABLE I
BREAK-DOWN ABLATION.

| Schemes | HFDI | MSDA | FA | MLRL | FCF | NUDT-SIRST (663: 664) | | | | SIRST (224:96) | | | | IRSTD-1k (800:201) | | | |
|---|---|---|---|---|---|---|---|---|---|---|---|---|---|---|---|---|---|
| | | | | | | IoU | nIoU | $P_d$ | $F_a$ | IoU | nIoU | $P_d$ | $F_a$ | IoU | nIoU | $P_d$ | $F_a$ |
| Base | ✗ | ✗ | ✗ | ✗ | ✗ | 0.867 | 0.887 | 0.976 | 17.37 | 0.742 | 0.727 | 0.980 | 34.97 | 0.661 | 0.651 | 0.913 | 14.12 |
| Base+HFDI | ✓ | ✗ | ✗ | ✗ | ✗ | 0.873 | 0.886 | 0.979 | 13.54 | 0.756 | 0.736 | 0.970 | 27.30 | 0.670 | 0.659 | 0.916 | 19.42 |
| Base+MSDA | ✗ | ✓ | ✗ | ✗ | ✗ | 0.921 | 0.926 | 0.987 | 4.41 | 0.790 | 0.757 | 0.990 | 1.15 | 0.704 | 0.684 | 0.929 | 30.86 |
| Base+FA | ✗ | ✗ | ✓ | ✗ | ✗ | 0.877 | 0.887 | 0.975 | 8.43 | 0.749 | 0.735 | 0.980 | 28.33 | 0.668 | 0.672 | 0.892 | **11.16** |
| Base+MLRL | ✗ | ✗ | ✗ | ✓ | ✗ | 0.915 | 0.920 | 0.981 | 4.02 | 0.774 | 0.752 | 0.990 | 11.44 | 0.688 | 0.675 | 0.916 | 20.04 |
| Base+FCF | ✗ | ✗ | ✗ | ✗ | ✓ | 0.888 | 0.898 | 0.974 | 10.39 | 0.760 | 0.748 | 0.980 | 27.66 | 0.665 | 0.671 | 0.919 | 13.36 |
| Our-w/o HFDI | ✗ | ✓ | ✓ | ✓ | ✓ | 0.935 | 0.936 | 0.984 | **1.63** | 0.787 | 0.762 | 0.990 | 6.04 | 0.710 | 0.688 | 0.923 | 25.43 |
| Our-w/o MSDA | ✓ | ✗ | ✓ | ✓ | ✓ | 0.918 | 0.924 | 0.983 | 8.64 | 0.782 | 0.754 | 0.990 | 4.93 | 0.693 | 0.662 | 0.906 | 22.77 |
| Our-w/o FA | ✓ | ✓ | ✗ | ✓ | ✓ | 0.935 | 0.937 | 0.987 | 3.79 | 0.794 | 0.763 | 0.980 | **0.80** | 0.704 | 0.691 | 0.939 | 27.44 |
| Our-w/o MLRL | ✓ | ✓ | ✓ | ✗ | ✓ | 0.928 | 0.933 | 0.987 | 3.63 | 0.786 | 0.761 | 0.971 | 9.10 | 0.693 | 0.685 | 0.933 | 20.82 |
| Our-w/o FCF | ✓ | ✓ | ✓ | ✓ | ✗ | 0.927 | 0.932 | 0.985 | 2.18 | 0.792 | 0.766 | 0.990 | 6.20 | 0.713 | 0.672 | **0.948** | 26.29 |
| Ours | ✓ | ✓ | ✓ | ✓ | ✓ | **0.938** | **0.941** | **0.992** | 3.70 | **0.800** | **0.775** | **1.000** | 5.01 | **0.719** | **0.692** | 0.943 | 11.39 |

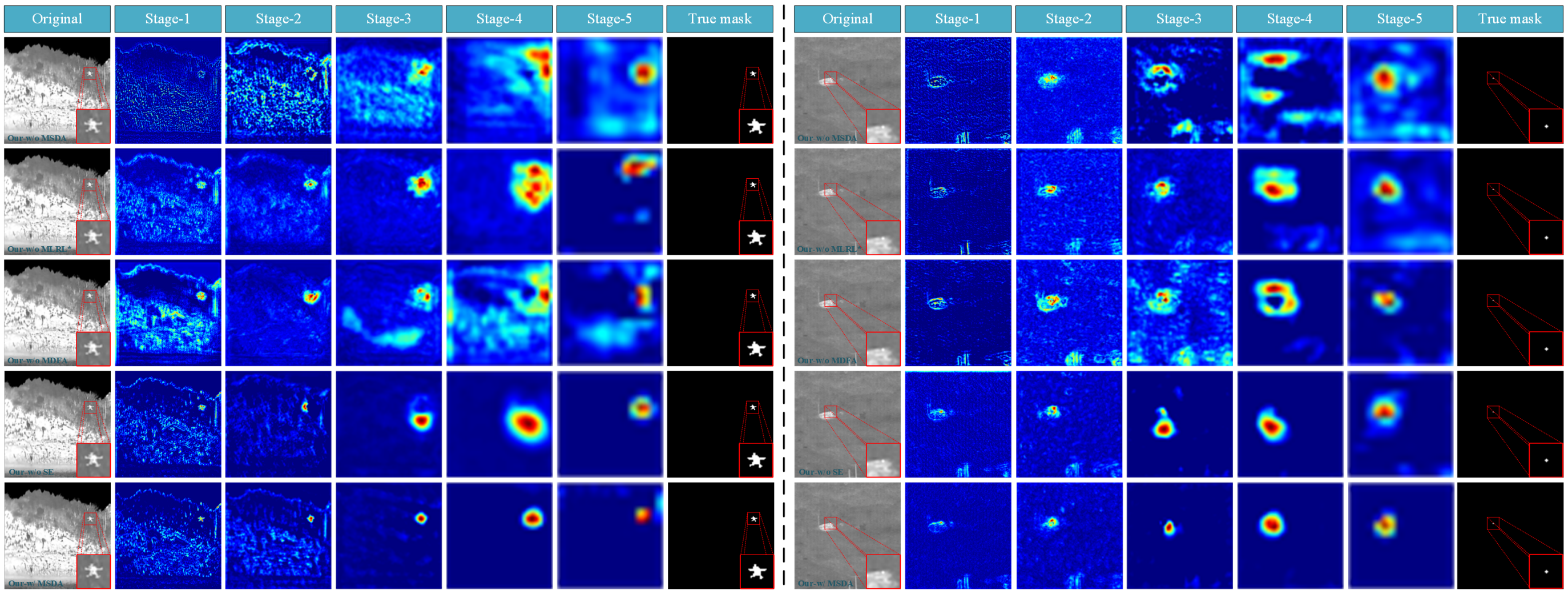

**Fig. 7.** Visualization of feature maps for the MSDA module and its component ablations to illustrate the contribution of each submodule to small target perception. From top to bottom: Our-w/o MSDA, Our-w/o MLRL*, Our-w/o MDFA, Our-w/o SE, Ours MSDA.

*2) SIRST dataset*. It is a real dataset with a total of 427 infrared images from different scenes. Approximately 10% of the images in this dataset contain multiple objects. For this dataset, we adopt two division rules: 341: 86 and 224: 96.

*3) IRSTD-1k dataset*. It is a real dataset consisting of 1001 infrared images of 512×512 pixels. This dataset contains small targets of different types and locations, such as drones, creatures, ships, and vehicles. At the same time, the dataset covers a variety of scenes, such as the sky, ocean, and land.

*4) IRSatVideo-LEO dataset*: It is a semi-simulated dataset that combines real satellite imagery with synthetic satellite motion, target appearances, trajectories, and intensities. It consists of 200 video sequences with a total of 91,366 frames, each annotated with precise pixel-level target masks. Following the official division rule [1], we use 160 sequences for training and 40 sequences for testing.

## *B. Experimental Settings*

*1) Experimental Environment and Parameter Settings*: The operating system is Ubuntu18.04.6, and the GPU is a RTX 2080Ti 11G. The batch size, learning rate and epochs are 4, $1e^{-4}$ and 500. The reduction ratio in the SE module is set to 4. Data augmentation operations such as random image flipping, rotation, and contrast enhancement are used.

*2) Evaluation Metrics*: Considering that the SIRST detection task is a segmentation task, we use mainly two pixel-level evaluation metrics: the intersection-over-union (IoU) metric and the normalized intersection-over-union (nIoU) metric. Their expressions are as follows:

$$IoU = \frac{\sum_{i=1}^{N} |P_i \cap T_i|}{\sum_{i=1}^{N} |P_i \cup T_i|} \tag{12}$$

$$nIoU = \frac{1}{N} \sum_{i=1}^{N} \frac{|P_i \cap T_i|}{|P_i \cup T_i|} \tag{13}$$

where $P_i$ denotes the set of predicted target pixels of the *i-th* image. $T_i$ denotes the true target pixel set of the *i-th* image.

TABLE II
ABLATION STUDY OF PROGRESSIVE MODULE INTEGRATION.

| Schemes | HFDI | MSDA | FA | MLRL | FCF | NUDT-SIRST (663: 664) | | | | SIRST (224:96) | | | | IRSTD-1k (800:201) | | | |
|---|---|---|---|---|---|---|---|---|---|---|---|---|---|---|---|---|---|
| | | | | | | IoU | nIoU | $P_d$ | $F_a$ | IoU | nIoU | $P_d$ | $F_a$ | IoU | nIoU | $P_d$ | $F_a$ |
| Base | ✗ | ✗ | ✗ | ✗ | ✗ | 0.867 | 0.887 | 0.976 | 17.37 | 0.742 | 0.727 | 0.980 | 34.97 | 0.661 | 0.651 | 0.913 | 14.12 |
| Scheme1 | ✗ | ✗ | ✗ | ✗ | ✓ | 0.888 | 0.898 | 0.974 | 10.39 | 0.760 | 0.748 | 0.980 | 27.66 | 0.665 | 0.671 | 0.919 | 13.36 |
| Scheme2 | ✗ | ✗ | ✗ | ✓ | ✓ | 0.909 | 0.921 | 0.984 | 13.40 | 0.777 | 0.749 | 0.971 | 6.56 | 0.682 | 0.678 | 0.923 | 13.42 |
| Scheme3 | ✗ | ✗ | ✓ | ✓ | ✓ | 0.913 | 0.919 | 0.984 | 6.99 | 0.780 | 0.753 | 0.990 | 10.09 | 0.690 | 0.678 | 0.909 | 22.98 |
| Scheme4 | ✗ | ✓ | ✓ | ✓ | ✓ | 0.935 | 0.936 | 0.984 | **1.63** | 0.787 | 0.762 | 0.990 | 6.04 | 0.710 | 0.688 | 0.923 | 25.43 |
| Ours | ✓ | ✓ | ✓ | ✓ | ✓ | **0.938** | **0.941** | **0.992** | 3.70 | **0.800** | **0.775** | **1.000** | **5.01** | **0.719** | **0.692** | **0.943** | **11.39** |

TABLE III
VERIFICATION OF EACH COMPONENT IN THE MSDA MODULE ON THE NUDT-SIRST DATASET.

| Schemes | MLRL* | MDFA | SE | IoU | nIoU | $P_d$ | $F_a$ |
|---|---|---|---|---|---|---|---|
| Our-w/o MSDA | ✗ | ✗ | ✗ | 0.918 | 0.924 | 0.983 | 8.64 |
| Our-w/o MLRL* | ✗ | ✓ | ✓ | 0.929 | 0.934 | 0.984 | **1.75** |
| Our-w/o MDFA | ✓ | ✗ | ✓ | 0.928 | 0.932 | 0.986 | 4.94 |
| Our-w/o SE | ✓ | ✓ | ✗ | 0.934 | 0.938 | 0.988 | 3.56 |
| Ours | ✓ | ✓ | ✓ | **0.938** | **0.941** | **0.992** | 3.70 |

TABLE IV
VERIFICATION OF EACH BRANCH IN THE MDFA MODULE ON THE NUDT-SIRST DATASET. LL, LH, HL, AND HH DENOTE THE BRANCHES CORRESPONDING TO LOW-LOW (LOW-FREQUENCY), LOW-HIGH (HORIZONTAL HIGH-FREQUENCY), HIGH-LOW (VERTICAL HIGH-FREQUENCY), AND HIGH-HIGH (DIAGONAL HIGH-FREQUENCY) COMPONENTS, RESPECTIVELY.

| Schemes | LL | LH | HL | HH | IoU | nIoU | $P_d$ | $F_a$ |
|---|---|---|---|---|---|---|---|---|
| Our-w/o MDFA | ✗ | ✗ | ✗ | ✗ | 0.928 | 0.932 | 0.986 | 4.94 |
| Our-w/o LL | ✗ | ✓ | ✓ | ✓ | 0.932 | 0.934 | 0.984 | 2.83 |
| Our-w/o LH | ✓ | ✗ | ✓ | ✓ | 0.932 | 0.935 | 0.987 | 2.83 |
| Our-w/o HL | ✓ | ✓ | ✗ | ✓ | 0.933 | 0.936 | 0.984 | **2.69** |
| Our-w/o HH | ✓ | ✓ | ✓ | ✗ | 0.930 | 0.936 | 0.984 | 3.42 |
| Ours | ✓ | ✓ | ✓ | ✓ | **0.938** | **0.941** | **0.992** | 3.70 |

$|P_i \cap T_i|$ denotes the number of pixels in the intersection of the predicted regions and the ground-truth regions in the *i-th* image. $|P_i \cup T_i|$ denotes the number of pixels in the union of the predicted regions and ground-truth regions in the *i-th* image. The IoU tends to reflect the model's detection ability for larger small targets, and the nIoU tends to reflect the model's detection ability for smaller targets.

At the same time, we add two target-level evaluation metrics, namely, the detection rate $P_d$ and the false alarm rate $F_a$. Their expressions are as follows:

$$P_d = \frac{T_{correct}}{T_{All}} \tag{14}$$

$$F_a = \frac{P_{false}}{P_{all}} \tag{15}$$

where $T_{correct}$ denotes the number of correctly predicted targets. $T_{All}$ denotes the true target number. $P_{false}$ denotes the number of incorrectly predicted target pixels. $P_{all}$ denotes the total number of pixels in the image. Specifically, when the centroid deviation of the target is less than a predefined deviation threshold, the target is considered to be correctly predicted. Otherwise, the prediction is considered incorrect. Consistent with [34], we set the predefined deviation threshold to 3. The default magnitude of $F_a$ is $10^{-6}$. In the presented experimental results, ***bold*** denotes the best result, and *underline* denotes the second best result.

*3) Loss Function:* To address the significant class imbalance between targets and background in the SIRST detection tasks, we adopt the SoftIoU loss [64] as the loss function. The formulation is as follows:

$$Loss_{softIoU}(p, y) = \frac{\sum_{i,j} p_{i,j} \cdot y_{i,j}}{\sum_{i,j} (p_{i,j} + y_{i,j} - p_{i,j} \cdot y_{i,j})} \tag{16}$$

where $p$ denotes the predicted score map output by the network, and $y$ denotes the corresponding ground-truth mask.

*4) Relative rate of change:* To compare the performance impact of the proposed method more clearly, we adopt the relative rate of change to quantify the improvement or degradation. The expression is as follows:

$$I_{out} = \frac{(R_{after} - R_{before})}{R_{before}} \times 100\% \tag{17}$$

where $R_{before}$ and $R_{after}$ denote the performance metrics of the initial state (e.g., the model without the module) and the changed state (e.g., the model with the module added), respectively.

### C. Ablation Experiment

To fully verify the performance of MSDA-Net, we conduct a large number of ablation experiments on multiple influencing factors on the NUDT-SIRST (663: 664) dataset, SIRST (224:96) dataset, and IRSTD-1k (800:201) dataset.

*1) Break-down ablation:* To fully verify the effect of each component, we conduct detailed ablation experiments on the HFDI module, MSDA module, FA structure, MLRL module, and FCF module. The experimental results on the three datasets can be seen in Table I. From the experimental results, the network performance decreases the most when the MSDA module is removed. Specifically, the IoU decreases by 2.13% (from 0.938 to 0.918) ~ 3.62% (from 0.719 to 0.693), and the nIoU decreases by 1.81% (from 0.941 to 0.924) ~ 4.34% (from 0.692 to 0.662). Similarly, the network performance improves the most when only the MSDA module is added. Specifically, the IoU increases by 6.23% (from 0.867 to 0.921)

TABLE V
COMPARISON OF MSDA-NET AND VARIOUS SOTA METHODS ON THE NUDT-SIRST DATASET.

| Methods | IoU | | nIoU | | $P_d$ | | $F_a$ | |
|---|---|---|---|---|---|---|---|---|
| | 663: 664 | 1027: 300 | 663: 664 | 1027: 300 | 663: 664 | 1027: 300 | 663: 664 | 1027: 300 |
| FKRW[15] (TGRS'19) | 0.110 | 0.116 | 0.232 | 0.241 | - | - | - | - |
| MPCM[21] (PR'16) | 0.123 | 0.110 | 0.198 | 0.147 | - | - | - | - |
| IPI[23] (TIP'13) | 0.403 | 0.352 | 0.497 | 0.457 | - | - | - | - |
| ALCNet[36] (TGRS'21) | 0.822 | 0.830 | 0.835 | 0.844 | 0.990 | 0.979 | 7.24 | 13.07 |
| MLCL-Net[37] (IPT'22) | 0.895 | 0.912 | 0.904 | 0.914 | 0.970 | 0.993 | 7.84 | 15.36 |
| ALCL-Net[38] (GRSL'22) | 0.909 | 0.924 | 0.921 | 0.924 | 0.986 | 0.990 | 7.47 | 3.61 |
| ISNet[39] (CVPR'22) | 0.743 | 0.764 | 0.769 | 0.784 | 0.966 | 0.977 | 23.69 | 22.74 |
| AGPCNet[33] (TAES'23) | 0.841 | 0.863 | 0.863 | 0.881 | 0.973 | 0.977 | 11.90 | 6.61 |
| DNA-Net[34] (TIP'23) | 0.850 | 0.866 | 0.856 | 0.866 | 0.980 | 0.991 | 5.63 | **1.48** |
| UIU-Net[35] (TIP'23) | 0.903 | 0.931 | 0.897 | 0.925 | 0.985 | 0.986 | 4.46 | 2.14 |
| MTU-Net[43] (TGRS 23) | 0.793 | 0.817 | 0.822 | 0.837 | 0.964 | 0.970 | 36.38 | 11.29 |
| GGL-Net[40] (GRSL'23) | 0.923 | 0.940 | 0.934 | 0.940 | 0.989 | 0.993 | 4.44 | 2.39 |
| MSHNet[41] (CVPR'24) | 0.781 | 0.795 | 0.807 | 0.817 | 0.964 | 0.981 | 23.9 | 29.86 |
| RPCANet[46] (WACV'24) | 0.875 | 0.902 | 0.891 | 0.924 | 0.956 | 0.967 | 17.33 | 26.70 |
| SCTransNet[44] (TGRS'24) | 0.937 | 0.948 | **0.942** | **0.951** | 0.988 | 0.993 | 4.99 | 3.51 |
| MMLNet[47] (TGRS'25) | 0.826 | 0.847 | 0.856 | 0.863 | 0.973 | 0.974 | 14.87 | 10.98 |
| SDSNet [45] (TGRS'25) | 0.935 | 0.948 | 0.939 | 0.950 | 0.988 | 0.993 | 10.78 | 4.22 |
| ILNet [48] (TAES'25) | 0.917 | 0.928 | 0.916 | 0.927 | 0.985 | 0.993 | **3.33** | 3.15 |
| **MSDA-Net (Ours)** | **0.938** | **0.950** | 0.941 | **0.951** | **0.992** | **0.995** | 3.70 | 3.66 |

TABLE VI
COMPARISON OF MSDA-NET AND VARIOUS SOTA METHODS ON THE SIRST DATASET.

| Methods | IoU | | nIoU | | $P_d$ | | $F_a$ | |
|---|---|---|---|---|---|---|---|---|
| | 341: 86 | 224:96 | 341: 86 | 224:96 | 341: 86 | 224:96 | 341: 86 | 224:96 |
| FKRW[15] (TGRS'19) | 0.125 | 0.143 | 0.208 | 0.229 | - | - | - | - |
| MPCM[21] (PR'16) | 0.257 | 0.160 | 0.367 | 0.299 | - | - | - | - |
| IPI[23] (TIP'13) | 0.259 | 0.130 | 0.346 | 0.224 | - | - | - | - |
| ALCNet[36] (TGRS'21) | 0.737 | 0.753 | 0.745 | 0.732 | 0.973 | 0.971 | 9.58 | 31.63 |
| MLCL-Net[37] (IPT'22) | 0.732 | 0.757 | 0.774 | 0.741 | 0.982 | 0.980 | 37.57 | 22.33 |
| ALCL-Net[38] (GRSL'22) | 0.787 | 0.782 | 0.772 | 0.758 | 0.991 | **1.0** | 16.77 | 9.50 |
| ISNet[39] (CVPR'22) | 0.762 | 0.753 | 0.762 | 0.717 | 0.991 | 0.961 | 34.64 | 20.98 |
| AGPCNet[33] (TAES'23) | 0.761 | 0.747 | 0.754 | 0.713 | 0.991 | 0.990 | **2.04** | 15.22 |
| DNA-Net[34] (TIP'23) | 0.778 | 0.776 | 0.761 | 0.745 | 0.991 | 0.980 | 9.14 | 6.40 |
| UIU-Net[35] (TIP'23) | 0.779 | 0.763 | 0.749 | 0.742 | 0.982 | 0.990 | 22.40 | 42.68 |
| MTU-Net[43] (TGRS'23) | 0.754 | 0.762 | 0.750 | 0.722 | 0.982 | 0.971 | 22.18 | 8.38 |
| GGL-Net[40] (GRSL'23) | 0.806 | 0.795 | 0.783 | 0.768 | **1.0** | 0.990 | 4.35 | 7.07 |
| MSHNet[41] (CVPR'24) | 0.745 | 0.768 | 0.734 | 0.738 | 0.982 | **1.0** | 2.13 | **2.38** |
| RPCANet[46] (WACV'24) | 0.690 | 0.670 | 0.702 | 0.683 | 0.982 | 0.971 | 50.92 | 34.65 |
| SCTransNet[44] (TGRS'24) | 0.773 | 0.777 | 0.789 | 0.772 | 0.982 | 0.990 | 43.52 | 8.91 |
| MMLNet[47] (TGRS'25) | 0.724 | 0.730 | 0.739 | 0.718 | 0.954 | 0.980 | 12.46 | 24.16 |
| SDSNet [45] (TGRS'25) | 0.798 | 0.779 | 0.776 | 0.747 | 0.982 | 0.971 | 15.66 | 12.04 |
| ILNet [48] (TAES'25) | 0.794 | 0.782 | 0.773 | 0.756 | **1.0** | 0.990 | 4.79 | 3.90 |
| **MSDA-Net (Ours)** | **0.811** | **0.800** | **0.794** | **0.775** | **1.0** | **1.0** | 7.19 | 5.01 |

~ 6.51% (from 0.661 to 0.704) and the nIoU increases by 4.13% (from 0.727 to 0.757) ~ 5.07% (from 0.651 to 0.684). This shows that the features of infrared small targets can be effectively obtained by extracting the relationships between local areas at different scales and perceiving key features in different directions. At the same time, when the HFDI module is removed, the IoU and nIoU of the network decrease by 0.32% (from 0.938 to 0.935) ~ 1.63% (from

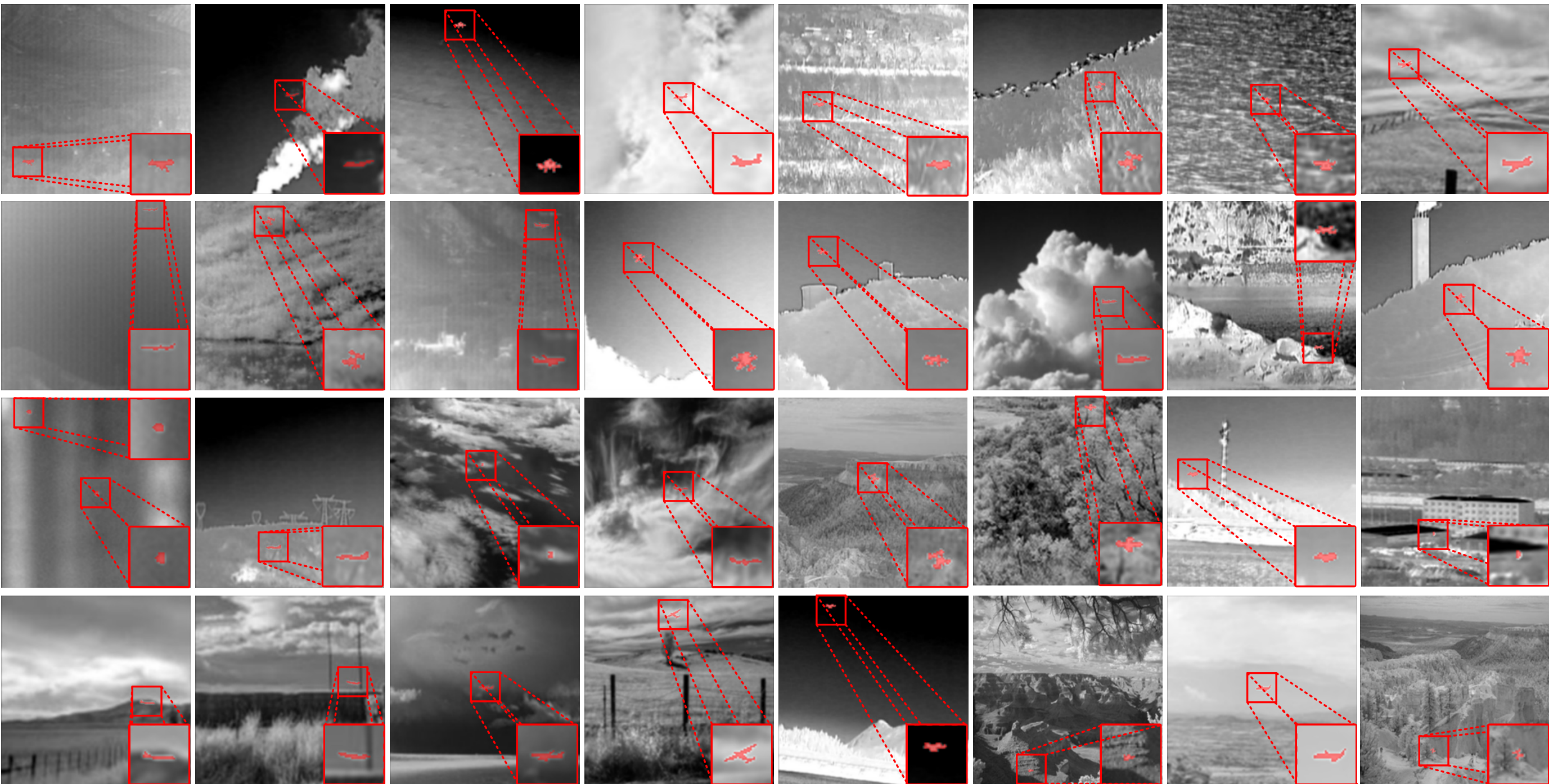

**Fig. 8.** Partial visualization results of the proposed MSDA-Net on the NUDT-SIRST dataset with a division rule of 663: 664.

0.800 to 0.787) and 0.53% (from 0.941 to 0.936) ~ 1.68% (from 0.775 to 0.762), respectively. This is because the proposed HFDI module helps emphasize detailed information such as textures, edges, and shapes that are relatively more important for the SIRST detection. When the FA structure is removed, the IoU and nIoU of the network decrease by 0.32% (from 0.938 to 0.935) ~ 2.09% (from 0.719 to 0.704) and 0.14% (from 0.692 to 0.691) ~ 1.55% (from 0.775 to 0.763), respectively. This is because the FA structure can effectively alleviate the problem of small targets disappearing due to network deepening. When the MLRL module is removed, the IoU and nIoU of the network decrease by 1.07% (from 0.938 to 0.928) ~ 3.62% (from 0.719 to 0.693) and 0.85% (from 0.941 to 0.933) ~ 1.81% (from 0.775 to 0.761), respectively. This is because extracting the relationships between local areas at different scales helps to fully mine the features of the target. When the FCF module is removed, the IoU and nIoU of the network decrease by 0.83% (from 0.719 to 0.713) ~ 1.17% (from 0.938 to 0.927) and 0.96% (from 0.941 to 0.932) ~ 2.89% (from 0.692 to 0.672), respectively. This is because the FCF module can effectively alleviate the target feature bias in multi-level feature map fusion.

*2) Progressive Module Ablation:* To further verify the rationality, complementarity, and overall effectiveness of each module, we conduct progressive module addition on three datasets. The results are presented in Table II. As each module is introduced sequentially, the model performance shows consistent improvement. Specifically, after adding the FCF module to the baseline, the IoU and nIoU on the three datasets improved by an average of 1.82% and 2.40%, respectively. This confirms the positive effect of the FCF module in alleviating feature bias during multi-scale feature fusion. With the further introduction of the MLRL module, the IoU and nIoU improved by an average of 2.39% and 1.24%, respectively, indicating that this module can effectively capture the relationships between local regions under varying receptive fields. With the addition of the FA module, the IoU and nIoU increased by an average of 0.67% and 0.10%, respectively. Although the improvement is relatively limited, this module effectively compensates for the lack of detailed information in high-level semantic features. Subsequently, after adding the MSDA module, the IoU and nIoU improved by an average of 2.07% and 1.51%, respectively, which demonstrates that extracting local relationships at different scales and perceiving key features in different directions can effectively mine the discriminative representation of infrared small targets. Finally, after the HFDI module was introduced, the IoU and nIoU increased by an average of 1.08% and 0.94%, respectively. This shows that the introduction of this module can enhance the expression of the edges and textures of small targets, providing more detailed information for subsequent feature extraction.

*3) Verification of each component in the MSDA module:* To further verify the effects of each component in the MSDA module, we conduct ablation experiments on the MLRL module, MDFA module and SE attention module in the MSDA module. To easily distinguish the results, we refer to the MLRL module in the MSDA module as the MLRL* module. According to Table III, both the MLRL* and MDFA modules can greatly improve network performance. At the same time, the performance improvement brought by the MDFA module is the greatest. Specifically, removing the MDFA module will cause the network to decrease the IoU and nIoU by 1.07% (from 0.938 to 0.928) and 0.96% (from 0.941 to 0.932), respectively. This is because MDFA can fully perceive key features in different directions, thereby guiding the network to refine the structure and position information of the target. At the same time, removing the MLRL* module decreases the IoU and nIoU of the network by 0.96% (from 0.938 to 0.929) and 0.74% (from 0.941 to 0.934), respectively. This finding verifies that the use of the MLRL* module in the feature extraction stage can achieve refined detection of small targets by promoting the extraction of relationships between local areas. In addition, removing the SE attention module also leads to performance degradation. This is because the SE attention module can effectively model the dependencies between different channels and guide the network to make full use of channel information to perform differentiated learning of the extracted features. To intuitively demonstrate the effectiveness of each component within the MSDA module, we present the feature maps after removing different sub-modules from the MSDA. As shown in Fig. 7, the complete MSDA module can significantly enhance the network's perception of small targets. Moreover, removing either MLRL* or MDFA leads to a degradation in the network's ability to extract critical features.

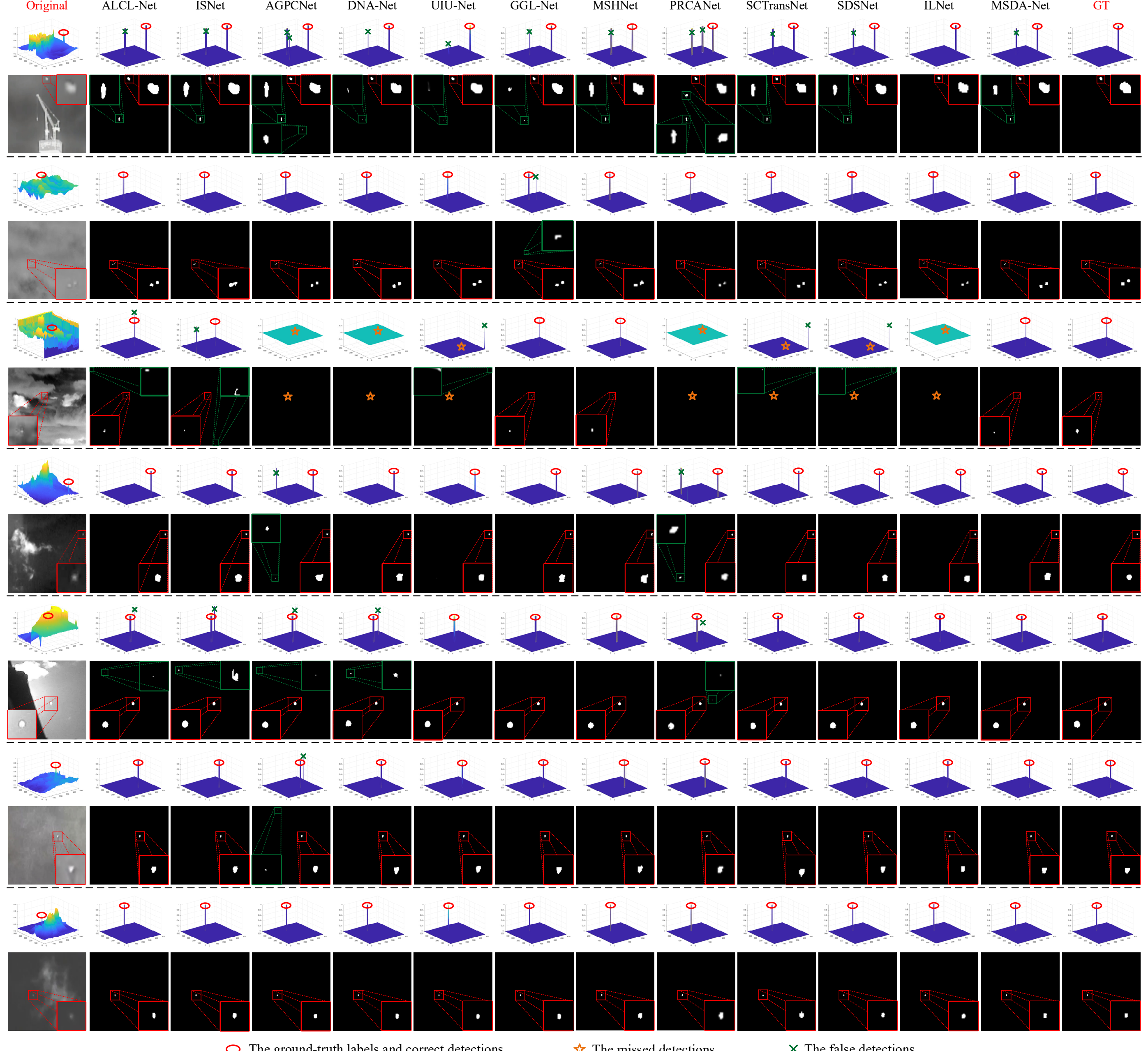

**Fig. 9.** Visualization results of various methods on difficult test samples of the SIRST dataset with a division rule of 224:96.

*4) Verification of each branch in the MDFA module:* To fully verify the effect of each branch of the MDFA module, we conduct detailed ablation experiments on each branch. From Table IV, removing a branch of the MDFA module will cause the IoU and nIoU to decrease. Specifically, removing the low-frequency branch (LL) will cause the network to decrease the IoU and nIoU by 0.64% (from 0.938 to 0.932) and 0.74% (from 0.941 to 0.934), respectively. This shows that the low-frequency branch plays an effective role in capturing global features and extracting high-level semantic information. In addition, removing the horizontal (LH), vertical (HL), or diagonal (HH) high-frequency branches causes the IoU and nIoU to decrease by an average of 0.64% (from 0.938 to 0.932) and 0.53% (from 0.941 to 0.936), respectively. This finding demonstrates that high-frequency directional information plays an important role in accurately capturing the edge and position information of small targets, and highlights the complementary nature of different directional branches in structural perception. In summary, each branch of the MDFA module contributes to the overall performance improvement. The low-frequency branch helps model global features, while the high-frequency branch helps enhance edge and texture representations in different directions. Preserving all branches allows the MDFA module to fully exploit multi-directional information and improve the model's robustness in detecting small targets under complex backgrounds.

TABLE VII
COMPARISON OF MSDA-NET AND VARIOUS SOTA METHODS ON THE IRSTD-1K DATASET.

| Scheme | IoU | nIoU | $P_d$ | $F_a$ | Parameters(M) |
|---|---|---|---|---|---|
| FKRW[15] | 0.092 | 0.159 | - | - | - |
| MPCM[21] | 0.055 | 0.215 | - | - | - |
| IPI[23] | 0.205 | 0.303 | - | - | - |
| ALCNet[36] | 0.666 | 0.660 | 0.929 | 10.17 | 0.38 |
| MLCL-Net[37] | 0.669 | 0.668 | 0.913 | 11.54 | 0.56 |
| ALCL-Net[38] | 0.707 | 0.669 | 0.943 | 13.11 | 5.67 |
| ISNet[39] | 0.653 | 0.631 | 0.916 | 34.43 | 0.65 |
| AGPCNet[33] | 0.670 | 0.651 | 0.906 | 13.61 | 12.36 |
| DNA-Net[34] | 0.684 | 0.618 | **0.956** | 15.11 | 4.70 |
| UIU-Net[35] | 0.664 | 0.652 | 0.899 | 25.05 | 50.54 |
| MTU-Net[43] | 0.674 | 0.618 | 0.879 | 31.18 | 12.75 |
| GGL-Net[40] | 0.683 | 0.681 | 0.939 | 30.37 | 8.99 |
| MSHNet[41] | 0.679 | 0.616 | 0.932 | 10.10 | 4.07 |
| RPCANet[46] | 0.615 | 0.603 | 0.876 | 10.93 | 0.68 |
| SCTransNet[44] | 0.670 | 0.668 | 0.923 | 18.16 | 11.33 |
| MMLNet[47] | 0.670 | 0.681 | 0.919 | 19.43 | 3.60 |
| SDSNet[45] | 0.677 | 0.676 | 0.933 | 12.49 | 2.55 |
| ILNet[48] | 0.683 | 0.686 | 0.936 | **4.08** | 4.04 |
| **MSDA-Net (Ours)** | **0.719** | **0.692** | 0.943 | 11.39 | 4.81 |

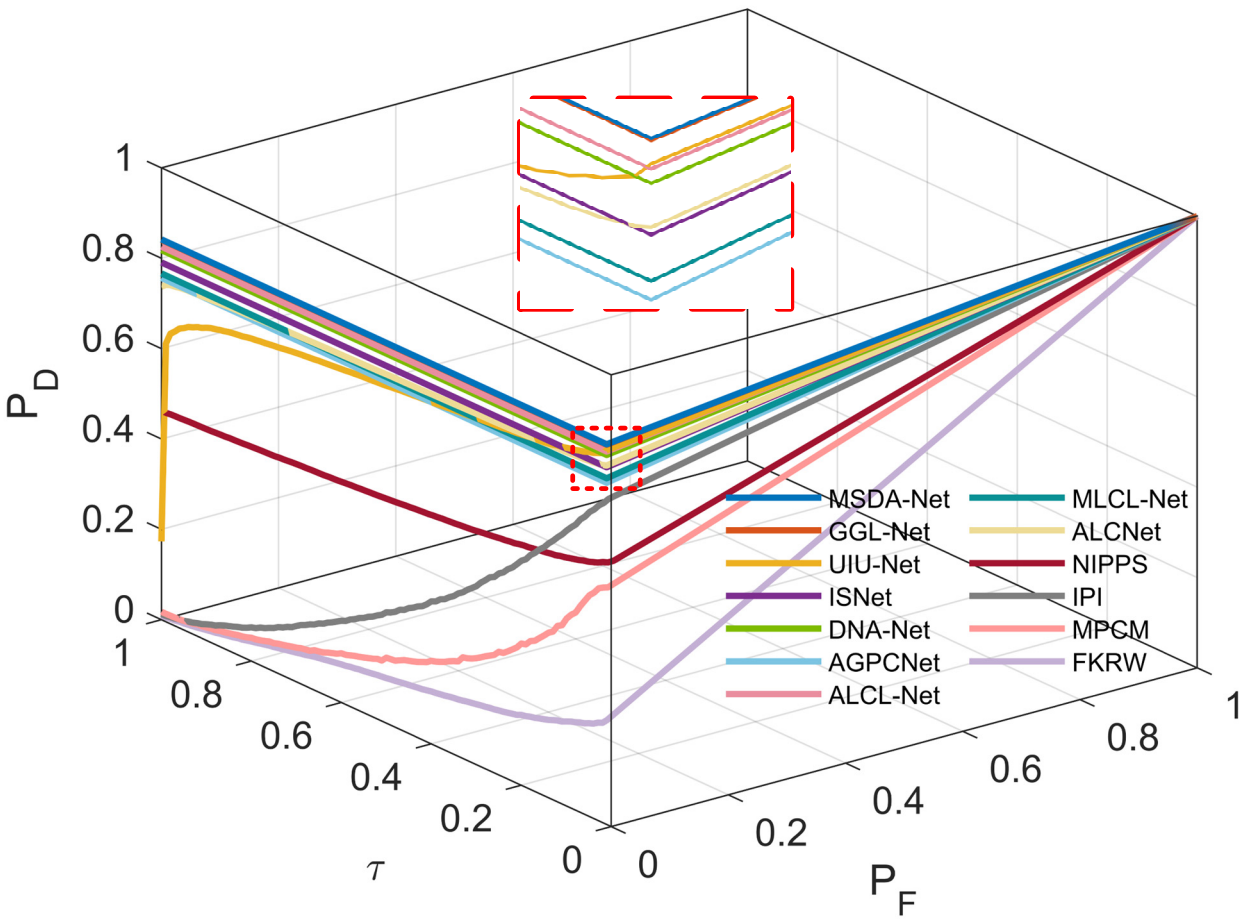

**Fig. 10.** 3D ROC curves using the IRSTD-1k dataset and uniform step size, △= 0.01. $P_D$ denotes the detection probability, $P_F$ denotes the false alarm probability, and τ denotes the threshold used by a detector.

*D. Comparison with other SOTA methods*

To fully prove the performance advantages of our MSDA-Net, we compare it with a variety of SOTA methods on three datasets: the NUDT-SIRST dataset, the SIRST dataset and the IRSTD-1k dataset. In addition, the hyper-parameter settings of the non-deep learning-based methods are consistent with those in [36].

*1) Performance comparison on the NUDT-SIRST dataset.* From Table V, we compare MSDA-Net with eighteen SOTA SIRST detection methods. Compared with non-deep learning-based methods, our proposed MSDA-Net achieves obvious performance improvements. This is because non-deep learning methods are usually designed for specific scenes, and their detection performance is poor for actual scenes with complex backgrounds and high interference. Compared with other deep learning-based methods, the proposed MSDA-Net achieves excellent results in both target-level and pixel-level evaluation metrics. Specifically, by averaging the experimental results under the two division rules of this dataset, MSDA-Net improves performance over other deep learning–based methods by 0.11% (from 0.943 to 0.944) ~ 25.20% (from 0.754 to 0.944) and 0.30% (from 0.991 to 0.994) ~ 3.33% (from 0.962 to 0.994) in terms of the IoU and $P_d$, respectively. Fig. 8 shows some visualization results of the MSDA-Net on the NUDT-SIRST dataset. It can be seen qualitatively that our MSDA-Net can accurately locate small targets and well segment the edge details of small targets with complex shapes. This is because our network pays attention to both the appearance features and high-frequency directional features of small targets, which is beneficial for achieving refined extraction of small targets. In addition, the proposed FCF module can also effectively alleviate the feature bias problem to further improve the segmentation accuracy of the network.

*2) Performance comparison on the SIRST dataset.* From Table VI, the deep learning-based methods achieve good results in terms of the target-level evaluation metrics $P_d$ and $F_a$. Notably, the $P_d$ of MSDA-Net reaches 1 for both dataset division rules of the SIRST dataset. This shows that MSDA-Net can detect all the targets in the test set. At the same time, our proposed MSDA-Net also achieves the best experimental results on the pixel-level evaluation metrics IoU and nIoU. Specifically, on the SIRST dataset with a division ratio of 341: 86, compared with other deep learning-based methods, MSDA-Net improves the performance by 0.62% (0.806 to 0.811) ~ 17.54% (from 0.690 to 0.811) in the IoU, and by 0.63% (from 0.789 to 0.794) ~ 13.11% (from 0.702 to 0.794) in the nIoU. On the SIRST dataset with a division ratio of 224: 96, compared with other deep learning-based methods, MSDA-Net improves the performance by 0.63% (from 0.795 to 0.800) ~ 19.40% (from 0.670 to 0.800) in IoU, and by 0.39% (from 0.772 to 0.775) ~ 13.47% (from 0.683 to 0.775) in nIoU. To evaluate the performance of each method more intuitively, we show the 2D and 3D detection results of different methods on the SIRST dataset. As shown in Fig. 9, although the proposed MSDA-Net still results in very few false detections on difficult samples, it can separate the target and background more effectively than other methods.

TABLE VIII
COMPARISON OF HAAR WAVELET OPERATORS AND OTHER MULTI-DIRECTIONAL HIGH-FREQUENCY OPERATORS

| Schemes | NUDT-SIRST (663: 664) | | | | SIRST (224:96) | | | | IRSTD-1k (800:201) | | | |
|---|---|---|---|---|---|---|---|---|---|---|---|---|
| | IoU | nIoU | $P_d$ | $F_a$ | IoU | nIoU | $P_d$ | $F_a$ | IoU | nIoU | $P_d$ | $F_a$ |
| Our-w/o HFDI | 0.935 | 0.937 | 0.984 | **1.63** | 0.787 | 0.762 | 0.990 | 6.04 | 0.710 | 0.688 | 0.923 | 25.43 |
| HFDI* | 0.936 | 0.940 | 0.989 | 4.23 | 0.798 | **0.777** | **1.000** | 7.03 | 0.712 | 0.689 | 0.939 | 31.01 |
| Our-w/o MSDA | 0.918 | 0.924 | 0.983 | 8.64 | 0.782 | 0.754 | 0.990 | **4.93** | 0.693 | 0.662 | 0.906 | 22.77 |
| MSDA* | 0.933 | 0.937 | 0.982 | 1.95 | 0.793 | 0.759 | 0.990 | 5.33 | 0.714 | 0.670 | 0.926 | 30.67 |
| Our (MSDA-Net) | **0.938** | **0.941** | **0.992** | 3.70 | **0.800** | 0.775 | **1.000** | 5.01 | **0.719** | **0.692** | **0.943** | **11.39** |

TABLE IX
EFFECTIVENESS OF HFDI AND MSDA FOR SMALLER TARGETS

| Scheme | NUDT-SIRST (663: 664) | | | | | | SIRST (224:96) | | | | | | IRSTD-1k (800:201) | | | | | |
|---|---|---|---|---|---|---|---|---|---|---|---|---|---|---|---|---|---|---|
| | Pixels≦9 | | Pixels≦16 | | ALL | | Pixels≦9 | | Pixels≦16 | | ALL | | Pixels≦9 | | Pixels≦16 | | ALL | |
| | IoU | Recall | IoU | Recall | IoU | Recall | IoU | Recall | IoU | Recall | IoU | Recall | IoU | Recall | IoU | Recall | IoU | Recall |
| Our-w/o HFDI | 0.953 | 0.965 | 0.933 | 0.947 | 0.935 | 0.961 | 0.588 | 0.833 | 0.587 | 0.788 | 0.787 | 0.892 | 0.537 | 0.806 | 0.613 | 0.816 | 0.710 | 0.836 |
| Our-w/o MSDA | 0.943 | 0.965 | 0.919 | 0.946 | 0.918 | 0.954 | 0.500 | 0.917 | 0.614 | 0.831 | 0.782 | 0.872 | 0.557 | 0.739 | 0.628 | 0.778 | 0.693 | 0.806 |
| Ours | **0.955** | **0.972** | **0.936** | **0.956** | **0.938** | **0.966** | **0.632** | **1.000** | **0.619** | **0.839** | **0.800** | **0.894** | **0.558** | **0.847** | **0.634** | **0.844** | **0.719** | **0.842** |

TABLE X
CROSS-DATASET GENERALIZATION PERFORMANCE: THE TWO SIDES OF THE ARROW DENOTE THE TRAINING AND TESTING DATASETS, RESPECTIVELY.

| Methods | SIRST → IRSTD-1k | | | | IRSTD-1k → SIRST | | | |
|---|---|---|---|---|---|---|---|---|
| | IoU | nIoU | $P_d$ | $F_a$ | IoU | nIoU | $P_d$ | $F_a$ |
| MLCL-Net | 0.403 | 0.483 | 0.872 | 228.02 | 0.520 | 0.615 | 0.817 | 105.79 |
| ALCL-Net | 0.350 | 0.489 | 0.896 | 331.48 | 0.595 | 0.643 | 0.826 | 43.96 |
| ISNet | 0.505 | 0.534 | 0.879 | 121.25 | 0.520 | 0.660 | 0.853 | 211.05 |
| AGPCNet | 0.495 | 0.501 | 0.872 | 127.69 | 0.578 | 0.654 | 0.908 | 33.70 |
| DNA-Net | 0.520 | 0.540 | 0.805 | 50.54 | 0.509 | 0.602 | 0.842 | 98.67 |
| UIU-Net | 0.352 | 0.445 | 0.862 | 292.02 | 0.545 | 0.579 | 0.899 | 21.25 |
| MTU-Net | 0.481 | 0.502 | 0.865 | 96.20 | 0.588 | 0.587 | 0.822 | 45.72 |
| GGL-Net | **0.536** | 0.552 | 0.872 | **28.35** | 0.555 | 0.690 | **0.945** | 180.22 |
| MSHNet | 0.523 | **0.568** | 0.852 | 31.88 | 0.476 | 0.569 | 0.807 | **17.03** |
| PRCA | 0.310 | 0.451 | 0.832 | 342.45 | 0.407 | 0.528 | 0.752 | 133.43 |
| SCTransNet | 0.463 | 0.548 | **0.899** | 63.34 | 0.620 | **0.716** | 0.927 | 38.23 |
| **MSDA-Net** | 0.527 | 0.552 | 0.875 | 101.27 | **0.639** | 0.702 | 0.927 | 17.97 |

*3) Performance comparison on the IRSTD-1k dataset.* Compared with the NUDT-SIRST and SIRST datasets, the IRSTD-1k dataset is more challenging due to the complexity and change of the scene. From Table VII, the proposed MSDA-Net achieves SOTA results on pixel-level evaluation metrics and excellent performance on target-level evaluation metrics. Specifically, on the IRSTD-1k dataset, compared with other deep learning-based methods, our MSDA-Net improves the IoU performance by 5.12% (from 0.684 to 0.719) ~ 16.91% (from 0.615 to 0.719) and the nIoU performance by 0.87% (from 0.686 to 0.692) ~ 14.76% (from 0.603 to 0.692). In addition, compared with those of UIU-Net, the proposed MSDA-Net parameters are reduced by 90.48%. To further evaluate the effectiveness of the proposed method, the 3D ROC curve [63] is used to evaluate the performance of various methods on the IRSTD-1k dataset. Compared with the traditional 2D ROC, the 3D ROC introduces a threshold parameter $\tau$ as the third dimension, forming a three-dimensional surface in the ($P_D$, $P_F$, $\tau$) space, thereby enabling a more comprehensive evaluation of the detector's performance across different threshold settings. From Fig. 10, the curve based on the deep learning method is significantly higher than the curve based on the non-deep learning method, which shows that the performance of the deep learning-based method is significantly better than that of the non-deep learning-based method. In addition, the dark blue curve of the proposed MSDA-Net is always at the highest position, which shows that it has the best target detection and background suppression performance. At the same time, we observe a very interesting phenomenon. For UIU-Net, the $P_D$ shows an obvious curve as the threshold changes. However, other deep learning-based methods result in smooth straight lines. The reason is that UIU-Net uses binary cross-entropy loss as the loss function, whereas other deep learning-based methods use softIoU loss [64]. Binary cross-entropy loss focuses on accurate segmentation of targets and background. Compared with softIoU loss, it takes more consideration into correctly predicting the background as background. This leads to more intermediate values between the target area and the background area in the UIU-Net prediction results.

## *E. Discussion*

*1) Rationality of the selected high-frequency operators.* To verify the superiority of using the high-frequency operator of the two-dimensional Haar wavelet transform to extract multi-directional high-frequency information in the task of infrared small target detection, we conduct comparative experiments between the high-frequency operator of the two-dimensional Haar wavelet transform and the multi-directional high-frequency operator with a kernel size of 3×3 on three datasets.

TABLE XI
PERFORMANCE COMPARISON BETWEEN HAAR WAVELET OPERATORS AND LEARNABLE CONVOLUTION KERNELS

| Schemes | NUDT-SIRST (663: 664) | | | | SIRST (224:96) | | | | IRSTD-1k (800:201) | | | |
|---|---|---|---|---|---|---|---|---|---|---|---|---|
| | IoU | nIoU | $P_d$ | $F_a$ | IoU | nIoU | $P_d$ | $F_a$ | IoU | nIoU | $P_d$ | $F_a$ |
| MDFA-Conv2 | 0.934 | 0.937 | 0.984 | 2.25 | 0.788 | 0.763 | 0.990 | 10.13 | 0.709 | 0.670 | 0.909 | 27.06 |
| MDFA-Conv3 | 0.935 | 0.937 | 0.984 | **1.28** | 0.793 | 0.766 | 0.971 | 9.74 | 0.710 | 0.678 | 0.919 | 25.94 |
| Our (MSDA-Net) | **0.938** | **0.941** | **0.992** | 3.70 | **0.800** | **0.775** | **1.000** | **5.01** | **0.719** | **0.692** | **0.943** | **11.39** |

TABLE XII
PERFORMANCE COMPARISON OF DIFFERENT SINGLE-FRAME DETECTORS INTEGRATED ON THE RFR FRAMEWORK ON THE IRSATVIDEO-LEO DATASET

| Methods | Easy | | | Medium | | | Complex | | | Total | | |
|---|---|---|---|---|---|---|---|---|---|---|---|---|
| | $P_d$ | $F_a$ | AUC | $P_d$ | $F_a$ | AUC | $P_d$ | $F_a$ | AUC | $P_d$ | $F_a$ | AUC |
| ALCNet_RFR | 0.974 | 12.44 | **0.995** | 0.766 | 6.56 | 0.895 | 0.725 | 5.44 | 0.923 | 0.825 | 8.23 | 0.936 |
| DNA-Net_RFR | 0.963 | **4.28** | 0.971 | 0.789 | **2.83** | 0.857 | 0.809 | **2.21** | 0.847 | 0.850 | **3.15** | 0.893 |
| ISTUD-Net_RFR | 0.991 | 19.86 | 0.986 | 0.857 | 12.80 | 0.902 | 0.839 | 16.66 | 0.895 | 0.897 | 16.22 | 0.928 |
| ResUnet _RFR | **0.995** | 22.42 | 0.989 | 0.889 | 12.49 | 0.924 | 0.852 | 17.31 | 0.903 | 0.915 | 17.13 | 0.941 |
| MSDANet_RFR | **0.995** | 6.07 | 0.982 | **0.895** | 3.58 | **0.946** | **0.873** | 8.46 | **0.944** | **0.923** | 5.73 | **0.958** |

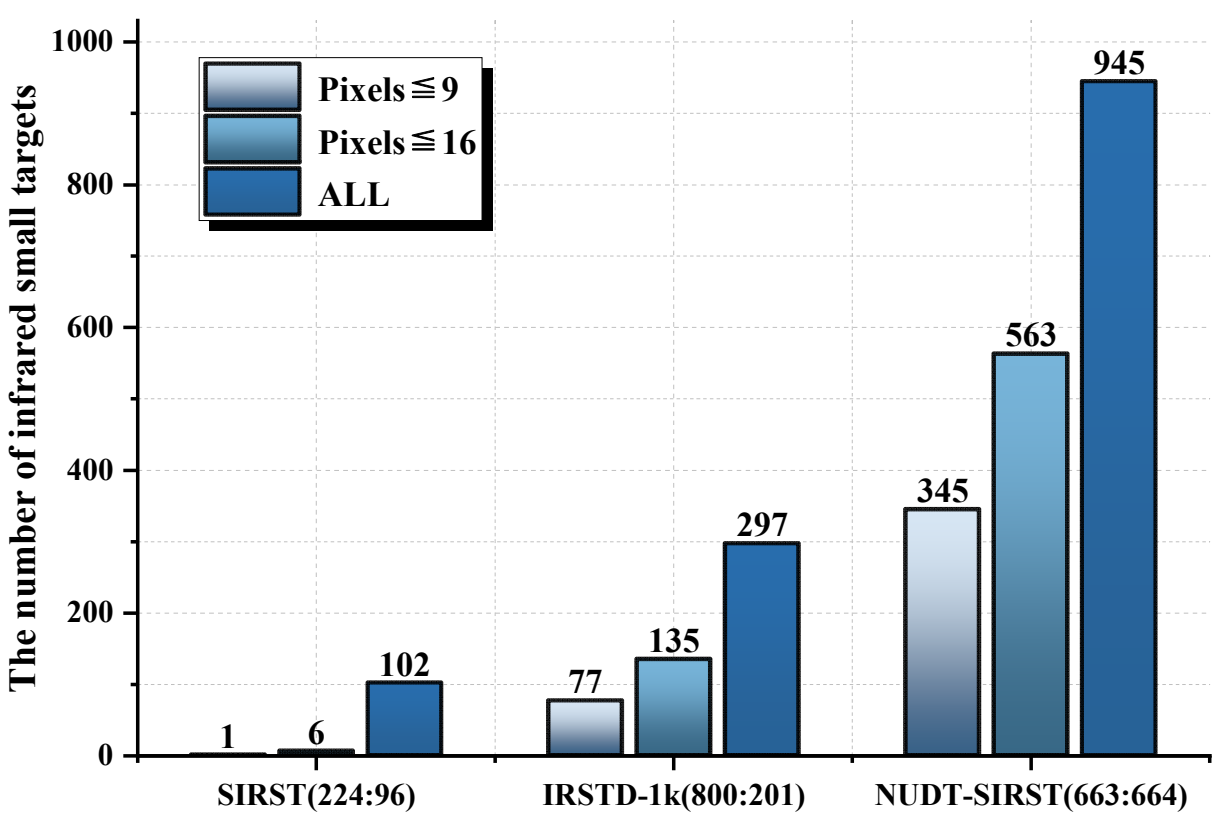

**Fig. 11.** Distribution statistics of target sizes in the test set on different datasets

Specifically, inspired by the Sobel operator, we construct a set of 3×3 multi-directional high-frequency operators: $\begin{bmatrix} 0 & -1 & -2 \\ 1 & 0 & -1 \\ 2 & 1 & 0 \end{bmatrix} \begin{bmatrix} -1 & 0 & 1 \\ -2 & 0 & 2 \\ -1 & 0 & 1 \end{bmatrix} \begin{bmatrix} -1 & -2 & -1 \\ 0 & 0 & 0 \\ 1 & 2 & 1 \end{bmatrix} \begin{bmatrix} -2 & -1 & 0 \\ -1 & 0 & 1 \\ 0 & 1 & 2 \end{bmatrix}$. The above operators are used to extract features in the horizontal direction, vertical direction, main diagonal direction and sub-diagonal direction. We replace the high-frequency operators of three two-dimensional Haar wavelet transforms in the HFDI and MSDA modules with this high-frequency operator, and conduct experiments on three public datasets. The HFDI and MSDA modules after the replacement of operators are called HFDI* and MSDA*, respectively. From Table VIII, the experimental results indicate that the use of the constructed 3×3 multi-directional high-frequency operator can still improve performance. However, compared with the high-frequency operator of the two-dimensional Haar wavelet transform used in MSDA-Net, the performance improvement it brings is limited. We analyse the possible reasons as follows: On the one hand, considering the small and weak characteristics of infrared small targets, the use of 3×3 multi-directional high-frequency operators may not be able to fully preserve the complex edges and subtle texture details in the image, especially for smaller infrared small targets. On the other hand, considering that the HFDI module is introduced into the second stage, interpolation is required when the 3×3 operator is used, which will cause further loss of detail information. Therefore, the high-frequency operator of the two-dimensional Haar wavelet transform used in the constructed MSDA-Net to enhance the target features has a more obvious performance advantage.

*2) Implications of the HFDI and MSDA modules for smaller targets.* To verify the effectiveness of the proposed HFDI and MSDA modules on smaller targets, we conduct experiments on three public datasets. Specifically, we define targets whose target area is less than or equal to 9 pixels or less than or equal to 16 pixels as smaller targets. We use the pixel-level IoU and pixel-level Recall to evaluate the detection effect. Fig. 11 shows the target size distribution of the test set in each dataset. The specific experimental results are shown in Table IX. On the three public datasets, removing the HFDI module or the MSDA module will cause the network's detection performance for smaller targets to deteriorate. Specifically, on the SIRST dataset with a partitioning rule of 224:96, for small targets with areas less than or equal to 9 pixels, after eliminating the HFDI module, the IoU and Recall decreased by 6.96% (from 0.632 to 0.588) and 16.7% (from 1.000 to 0.833), respectively. At the same time, after eliminating the MSDA module on the final MSDA-Net, the IoU and Recall decreased by 20.89% (from 0.632 to 0.500) and 8.30% (from 1.000 to 0.917), respectively. The experimental results fully verify the effectiveness of using multi-directional high-frequency information to enhance target features for smaller targets in various scenarios. We analyse the reasons as follows: First, unlike existing deep learning-based methods (such as ISNet and GGL-Net) that focus on the edge and shape features of small targets, the proposed MSDA-Net focuses on multi-directional high-frequency information, which contains more additional detail information, such as the target texture. High-frequency directional features are sensitive to local changes and directional information, which helps improve the network's ability to capture smaller targets or very subtle image details. Secondly, for smaller targets, they are similar to "noise" in the image. Such targets appear as significant high-frequency information in the image. Therefore, the proposed MSDA-Net constructed directly from the perspective of fully extracting high-frequency directional features will be effective. In addition, we use the high-frequency operator of the two-dimensional Haar wavelet transform to extract high-frequency directional information. Smaller convolution kernels are more sensitive to local pixel changes so that smaller targets can be extracted more finely. The multi-directional high-frequency operator with a kernel size of 2×2 can effectively prevent subtle edges or texture details from being smoothed and maintain detailed information.

*3) Generalization discussion.* To comprehensively evaluate the generalization ability of the proposed MSDA-Net on unseen datasets, we conduct cross-dataset experiments on two publicly available real-world datasets: SIRST (341:86) and IRSTD-1k (800:201), and compare the results with several representative methods. Specifically, the model is trained on the training set of one dataset and directly tested on the test set of another. According to the Table X, despite the significant differences between the two datasets in terms of imaging sources, target types, and background complexity, MSDA-Net still demonstrates relatively stable and superior generalization performance. Specifically, when tested on the SIRST dataset, the cross-domain model achieves 78.79% (0.639/0.811) and 92.70% (0.927/1.000) of the IoU and $P_d$ performance of the model trained on the in-domain data, respectively. When tested on the IRSTD-1k dataset, the cross-domain achieves 73.30% (0.527/0.719) and 92.79% (0.875/0.943) of the IoU and $P_d$ performance of the model trained on the in-domain data, respectively. These results indicate that the proposed MSDA-Net does not overly rely on specific data distributions. Instead, by effectively exploiting high-frequency directional features and capturing multi-scale information, it extracts discriminative and generalizable features, thereby demonstrating strong cross-domain generalization capability.

*4) Comparison of Haar wavelet operators and learnable convolutions kernels.* To better evaluate the performance differences between using fixed Haar wavelet operators and learnable convolution kernels in MDFA, we conduct comparative experiments on multiple public datasets. Specifically, we compare the fixed Haar wavelet operators with two learnable convolution kernel schemes: 1. MDFA-Conv2: In this variant, the wavelet operators of each directional branch in MDFA are replaced with 2×2 learnable convolution kernels. 2. MDFA-Conv3: In this variant, the wavelet operators of each directional branch in MDFA are replaced with 3×3 learnable convolution kernels. The results in Table XI show that compared with the above-mentioned learnable convolution kernel schemes, the fixed Haar wavelet operator achieves better performance. Specifically, in this task, the Haar wavelet operator provides a stable and direct method for extracting multi-directional high-frequency features. By explicitly focusing on directional information, it enhances the model's capability to perceive small-scale structures and weak-contrast edges, thereby ensuring robustness and stability of the model in complex scenarios. Meanwhile, SIRST tasks are often constrained by the small size of available datasets and strong noise interference. The fixed Haar wavelet operator offers a stable prior constraint, thereby enhancing the model's generalization and consistency. In addition, MSDA-Net has 4.81M parameters, while the parameter counts of the MDFA-Conv2 and MDFA-Conv3 schemes increase to 5.24M and 5.78M, respectively, yet their performance does not exceed that of the proposed method. Therefore, adopting the Haar wavelet operator in this method has obvious advantages.

### *F. Application*

In practical deployment scenarios, infrared targets often appear in continuous video streams or temporal image sequences. To evaluate the proposed method's temporal consistency and sustained detection performance in multi-frame conditions, we integrate MSDA-Net into the RFR [1], which is a recently introduced general multi-frame detection framework. This framework can flexibly integrate SIRST detection networks and enhance temporal target perception and retention capabilities through a recursive mechanism, thereby enabling multi-frame detection. We conduct experiments on the IRSATVideo-LEO dataset, and the detailed results are presented in Table XII. Compared with other methods, MSDA-Net equipped with the RFR framework achieves improvements in $P_d$ by 0.87% (from 0.915 to 0.923) ~ 11.88% (from 0.825 to 0.923), and in AUC by 1.81% (from 0.941 to 0.958) ~ 7.28% (from 0.893 to 0.958). The experimental results indicate that although MSDA-Net is designed as a single-frame detector based on static images, its outstanding performance within the multi-frame sequential detection framework demonstrates strong potential for temporal generalization.

## VI. CONCLUSION

To fully exploit and utilize the high-frequency directional features of infrared small targets, we propose an innovative multi-scale direction-aware network (MSDA-Net), which is an end-to-end network based on data-driven and model-driven methods. First, a HFDI module is proposed, which uses the frequency domain knowledge of the infrared small target image to replace part of the neural network layer and injects the high-frequency directional information into the network. Secondly, an innovative MDFA module is proposed to emphasize the focus on high-frequency directional features. At the same time, based on this, we further build a MSDA module combined with the MLRL module. The MSDA module can promote the full extraction of local relations at different scales and the full perception of key features in different directions. Thirdly, in view of the small and weak characteristics of infrared small targets, we propose a FA structure that aggregates multi-level features to solve the problem of small targets disappearing in deep feature maps. Finally, to alleviate the target feature bias in multi-level feature fusion, we propose a FCF module. This module performs feature-domain calibration before fusion, thereby achieving precise cross-layer feature fusion. Extensive experiments on multiple public datasets demonstrate that our proposed MSDA-Net achieves SOTA results and strong generalizability.

In future research, we aim to apply Spiking Neural Networks (SNNs) [65-70] to multi-frame infrared small target detection tasks. Owing to their strengths in temporal modeling, low energy consumption, and interpretability, SNNs are well-suited to address challenges such as weak signals, small-scale, and fast-moving targets. Therefore, we plan to build an efficient, energy-efficient, and robust multi-frame detection framework.

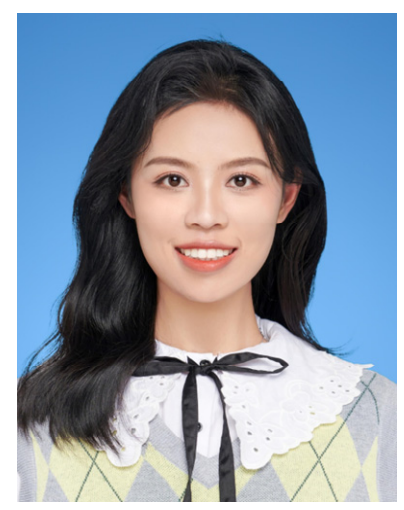

**Jinmiao Zhao** received the B.S. degree in Computer Science and Technology from the Northeast Forestry University, Harbin, China, in 2020. She is currently working toward the Ph.D. degree in Pattern Recognition and Intelligent Systems with the Shenyang Institute of Automation, Chinese Academy of Sciences, Shenyang, China.

Her current research interests include infrared small target detection and cross-spectral image patch matching.

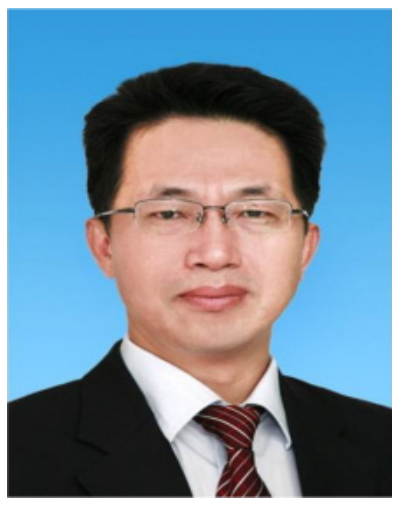

**Zelin Shi** received the B.S. and M.S. degrees from Xidian University, Xi'an, China, in 1987 and 1990, respectively. He received the Ph.D. degree in Mechatronics Engineering from the Shenyang Institute of Automation, Chinese Academy of Sciences, Shenyang, China, in 2005.

He is currently the Director of the Shenyang Institute of Automation, Chinese Academy of Sciences, and a Ph.D. supervisor. His research interests include opto-electronic imaging, image processing, pattern recognition, intelligent control, etc.

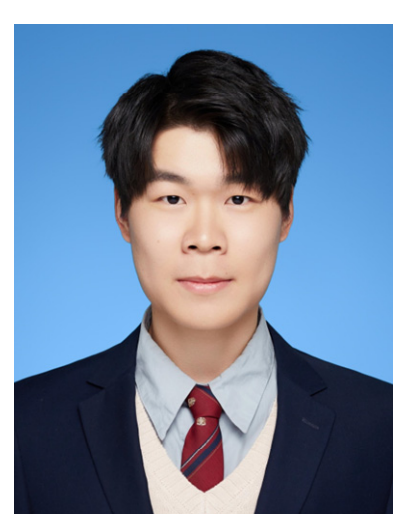

**Chuang Yu** (Graduate Student Member, IEEE) received the B.S. degree in Network Engineering from the Hainan University, Haikou, China, in 2020. He is working toward the Ph.D. degree in Pattern Recognition and Intelligent Systems with the Shenyang Institute of Automation, Chinese Academy of Sciences, Shenyang, China.

He is currently a research assistant at MMLab, The Chinese University of Hong Kong. His current research interests include cross-spectral image patch matching, infrared small target detection, and image segmentation.

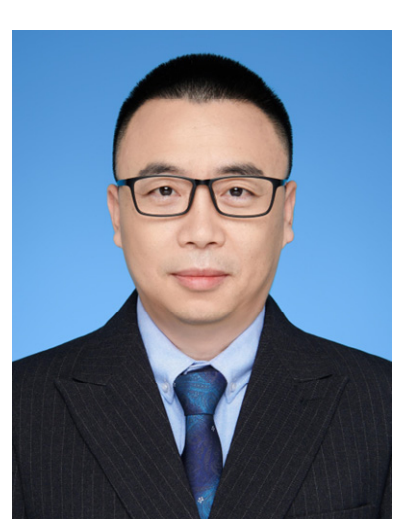

**Yunpeng Liu** (Member, IEEE) received the Ph.D. degree in Pattern Recognition and Machine Intelligence from the Shenyang Institute of Automation, Chinese Academy of Sciences, Shenyang, China, in 2010.

He is currently a Professor at the Shenyang Institute of Automation, Chinese Academy of Sciences, Shenyang, China. His current research interests include cross-spectral image patch matching, image segmentation, infrared small target detection, small target tracking, and recognition based on the Riemannian manifold.

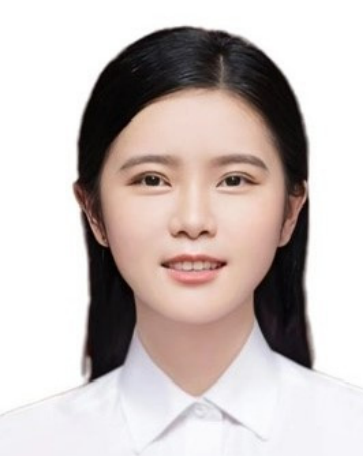

**Xinyi Ying** received the M.E. degree in Information and Communication Engineering from National University of Defense Technology (NUDT), Changsha, China, in 2021. She is currently pursuing the Ph.D. degree with the College of Electronic Science and Technology, NUDT.

Her current research interests focus on detection and tracking of infrared small targets.

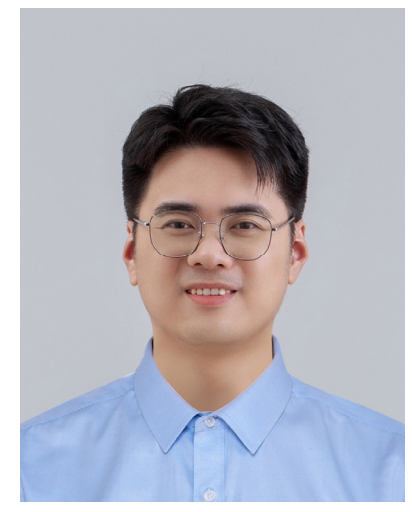

**Yimian Dai** (Member, IEEE) received the Ph.D. degree in Signal and Information Processing from the Nanjing University of Aeronautics and Astronautics, Nanjing, China, in 2021. He is currently an Associate Professor at the Visual Computing and Intelligent Perception (VCIP) Lab, College of Computer Science, Nankai University.

His current research interests include remote sensing object detection, visual perception in adverse environments, and efficient deployment and optimization within domestic AI ecosystem.